\documentclass[10pt,twocolumn,letterpaper,table]{article}

\usepackage[american]{babel}

\usepackage{multirow}
\usepackage{bbding}
\usepackage{tabularx}
\usepackage{float}

\usepackage{iccv} 

\definecolor{iccvblue}{rgb}{0.21,0.49,0.74}
\usepackage[pagebackref,breaklinks,colorlinks,allcolors=iccvblue]{hyperref}

\definecolor{cellpink}{rgb}{.99,.93,.98}

\newcommand{\stdvu}[1]{\scriptsize{\color{darkgray}(#1{\color{ForestGreen}$\uparrow$})} }
\newcommand{\stdvd}[1]{\scriptsize{\color{darkgray}(#1{\color{red}$\downarrow$})} }

\def \robotlogo {\raisebox{-0.1\height}{\includegraphics[height=1.2\baselineskip]{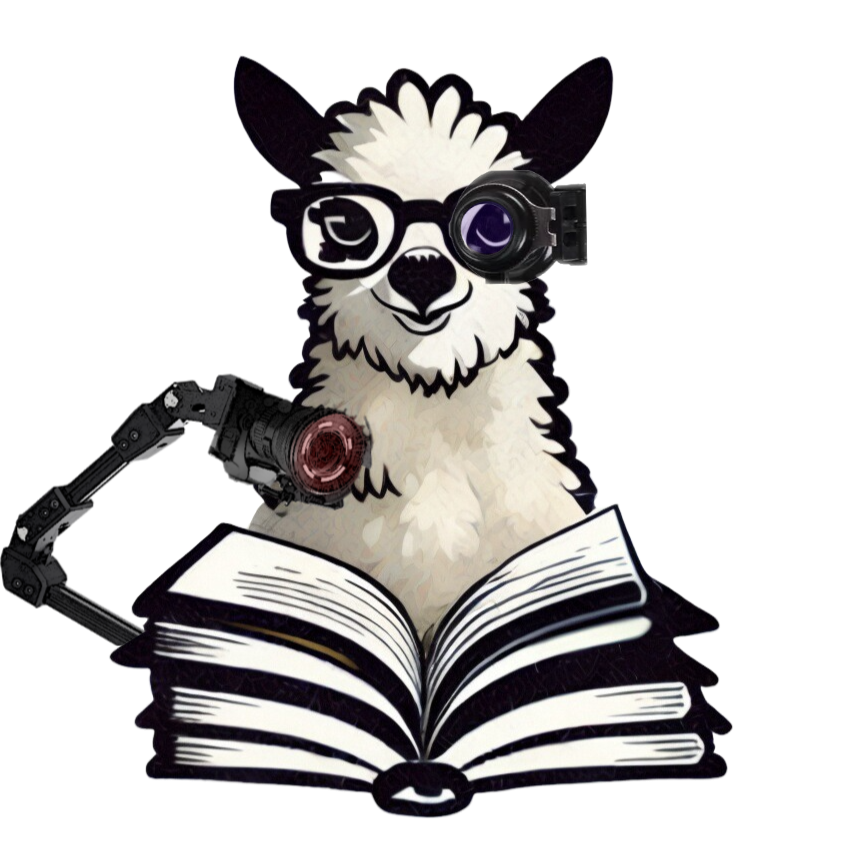}}}
\def\paperID{7845} 
\def\confName{ICCV}
\def\confYear{2025}

\title{\robotlogo{} Triad: Empowering LMM-based Anomaly Detection with\\Expert-guided Region-of-Interest Tokenizer and Manufacturing Process}

\author{
Yuanze Li$^{1\dagger}$ \quad Shihao Yuan$^{1,2\dagger}$ \quad Haolin Wang$^{1}$ \quad Qizhang Li$^{1,2}$\\ Ming Liu$^{1(}$\Envelope$^)$ \quad Chen Xu$^{2}$ \quad Guangming Shi$^{2}$ \quad Wangmeng Zuo$^{1,2}$\\
\tt\small{sqlyz@hit.edu.cn, csshihao@outlook.com, why\_cs@outlook.com, csqizhang@gmail.com}\\
\tt\small{csmliu@outlook.com, xc.xc@qq.com, gmshi@xidian.edu.cn, wmzuo@hit.edu.cn}\\
\small{$^{1}$Harbin Institute of Technology, $^{2}$Pengcheng Lab, Guangzhou}
}

\begin{document}
\maketitle
\begin{abstract}
Although recent methods have tried to introduce large multimodal models (LMMs) into industrial anomaly detection (IAD), their generalization in the IAD field is far inferior to that for general purposes.
We summarize the main reasons for this gap into two aspects.
On one hand, general-purpose LMMs lack cognition of defects in the visual modality, thereby failing to sufficiently focus on defect areas.
Therefore, we propose to modify the AnyRes structure of the LLaVA model, providing the potential anomalous areas identified by existing IAD models to the LMMs.
On the other hand, existing methods mainly focus on identifying defects by learning defect patterns or comparing with normal samples, yet they fall short of understanding the causes of these defects.
Considering that the generation of defects is closely related to the manufacturing process, we propose a manufacturing-driven IAD paradigm.
An instruction-tuning dataset for IAD (InstructIAD) and a data organization approach for Chain-of-Thought with manufacturing (CoT-M) are designed to leverage the manufacturing process for IAD.
Based on the above two modifications, we present \textbf{Triad}, a novel LMM-based method incorporating an expert-guided region-of-interest \textbf{t}okenizer and manufactu\textbf{r}ing process for \textbf{i}ndustrial \textbf{a}nomaly \textbf{d}etection.
Extensive experiments show that our Triad not only demonstrates competitive performance against current LMMs but also achieves further improved accuracy when equipped with manufacturing processes.
Source code, training data, and pre-trained models will be publicly available at \url{https://github.com/tzjtatata/Triad}.

\end{abstract}
\section{Introduction}
\label{sec:intro}

\begin{figure*}
\centering
\includegraphics[width=\linewidth]{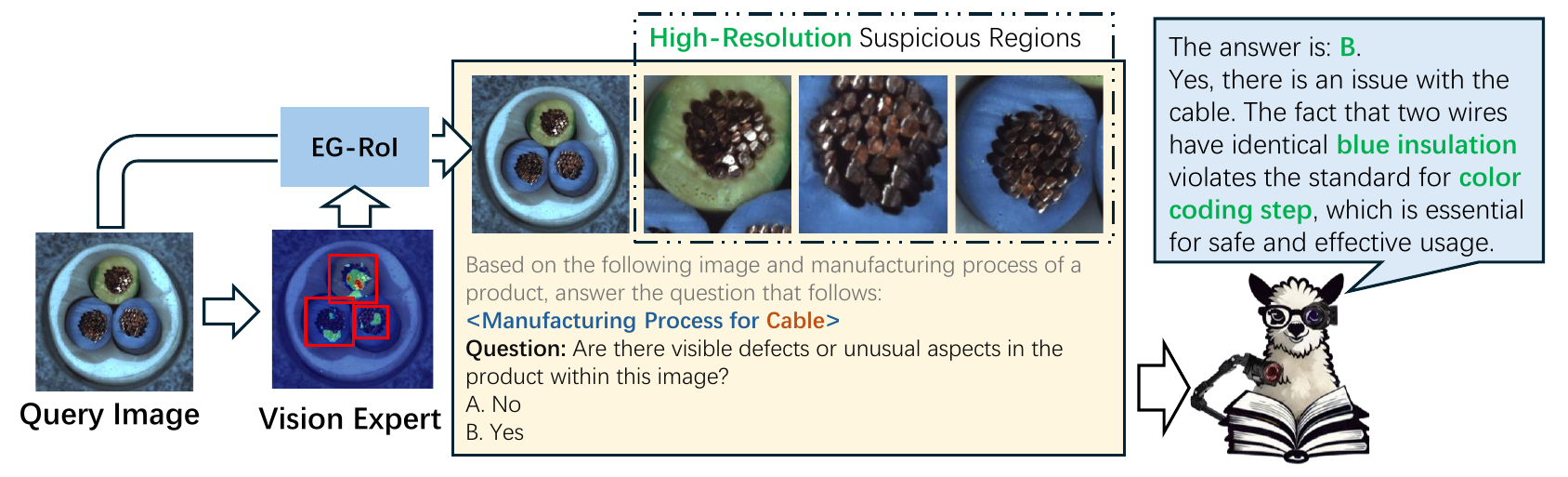}
\caption{Main workflow of Triad. As shown in the figure, the query image is first passed through EG-RoI with the suspicious regions predicted by the vision expert. These regions are cropped to keep a higher resolution than the query image and then encoded along with it. The text input includes a basic prompt, the manufacturing process for the specific product~(this kind of cable for this case) and a question about anomaly detection. In this case, Triad finds out that the issue is the wrong insulation color, as the manufacturing process mentioned that the cable is composed of three different color wires.}
\label{fig: overview}\vskip -0.2in
\end{figure*}

Industrial Anomaly Detection (IAD) plays a crucial role in modern manufacturing, where continuous iteration of product design and diverse inspection criteria call for robust and adaptable inspection solutions. Although large multimodal models (LMMs) have recently shown remarkable success on general-purpose tasks, their performance in IAD remains noticeably below expectations. We attribute this discrepancy to two key factors.

First, current LMMs are typically trained to align vision and language modalities for broad semantic understanding, yet they lack the specialized ability to identify and focus on potential defect regions. Industrial defects often appear subtly against complex backgrounds or in close proximity to functional components, making a purely global, coarse alignment insufficient.
To address this challenge, we propose an expert-guided region-of-interest tokenizer~(EG-RoI). Inspired by AnyRes module in LLaVA~\cite{llava, llavanext, llavaov}, EG-RoI extracts high-resolution and potentially anomalous regions of interest identified by existing IAD methods. By explicitly highlighting anomalous and normal regions during training, we enable the model to learn targeted visual cues indicative of attributes of products and defects. During inference, the high resolution suspicious regions predicted by vision experts further serve as key references for more precise anomaly detection.

Second, prevailing IAD approaches treat defects as isolated visual phenomena, neglecting their intrinsic relationship to manufacturing~(MFG) workflows. In reality, anomalies arise from process deviations (material impurities, assembly errors, or equipment malfunctions) that propagate through production stages. Existing methods, which rely on image-level comparisons or defect taxonomies, fail to leverage this causal knowledge.
To bridge this gap, we introduce a manufacturing-driven industry anomaly detection paradigm that weaves relevant manufacturing processes into the reasoning process. Our framework incorporates: (1)~An instruction-tuning dataset with attribute-rich captions, InstructIAD, spanning a variety of products and defect types from existing IAD datasets; (2) Chain-of-Thought with Manufacturing (CoT-M), a data organization strategy that synthesizes defect causality by linking anomalies to specific manufacturing steps. CoT-M expands InstructIAD by (i) editing normal product captions at the attribute level to simulate product evolution and defect occurrence and (ii) generating Chain-of-Thought processes grounded in manufacturing steps via GPT. When captions are unavailable, CoT-M employs a checklist-style template mirroring human inspector workflows.
Ultimately, we propose \textbf{Triad}, a novel LMM-based method incorporating an expert-guided region-of-interest \textbf{t}okenizer and manufactu\textbf{r}ing process for \textbf{i}ndustrial \textbf{a}nomaly \textbf{d}etection. 
We evaluate Triad on three benchmarks, MVTec-AD~\cite{bergmann2019mvtec}, WFDD~\cite{wfdd}, and PCB-Bank~\cite{pcb_bank}, and assess results under 0-/1-shot settings. Experiments show that Triad-ov-7B not only achieves competitive performance against both general-purpose and domain-specific LMMs but also demonstrates further improvements when accounting for manufacturing processes. Notably, with a single reference image (1-shot), Triad-ov-7B achieves 94.1\% on MVTec-AD, surpassing Qwen2-VL-72B~\cite{qwen2vl} by 2.1\%. Extensive analyses show the ability of Triad to comprehend complex manufacturing workflows and extend to unseen processes.

In summary, the contributions of this paper include,
\begin{itemize}

\item Triad, a novel LMM equipped with an expert-guided region-of-interest tokenizer and manufacturing-aware Chain-of-Thought reasoning, achieving state-of-the-art 0-/1-shot anomaly detection performance.

\item A data organization strategy CoT-M and a human-annotated instruction-tuning dataset InstructIAD, augmenting LMMs with causal defect understanding.

\item Comprehensive zero-/one-shot benchmarks showing Triad’s superiority over general-purpose and domain-specific LMMs, with analyses demonstrating its ability to generalize to unseen manufacturing processes.

\end{itemize}
\section{Related Works}
\label{sec:related}
\subsection{Industrial Anomaly Dectection}

Traditional unsupervised IAD methods typically fall into two categories. Reconstruction-based methods~\cite{draem, omni, realnet, pcb_bank} regenerate defect-free images to highlight deviations, while feature-embedding methods~\cite{patchcore, revisiting, simplenet, pyramidflow} compare query embeddings against normal feature patterns. Although effective, these methods usually require a dedicated model for each product, limiting their adaptability in dynamic production settings.

Recent works have leveraged vision-language models for zero-shot and few-shot anomaly detection by aligning visual features with textual prompts. For instance, WinClip~\cite{jeong2023winclip} measures similarity using hand-crafted text prompts, and later methods~\cite{chen2023april, clipad} enhance these prompts with visual cues or learned representations. However, the limited reasoning capabilities of their base models still present challenges. 
Our work addresses this gap by introducing long CoT answers during training to fully harness the reasoning power of large multimodal models.
\subsection{Large Multimodal Models for IAD}

Recent progress in LMMs has led to significant advances across a variety of vision-language tasks, driven by improvements in visual-encoder architectures such as Q-Former~\cite{li2023blip}, adaptive visual encoding~\cite{minicpm_v}, and spatial-aware visual sampling~\cite{ferret2}. Several IAD-centric methods attempt to incorporate such models by supplying textual guidance. For instance, Customizable-VLM~\cite{cvlm} tailors off-the-shelf LMMs to anomaly detection using class-specific prompts and normality descriptions, while MMAD~\cite{jiang2024mmad} employs a defect-aware strategy to integrate a database of normal product features and defect descriptions. Although these textual cues enhance performance, they still fall short of capturing the underlying interactions among manufacturing steps that directly lead to defects.

The most closely related efforts include AnomalyGPT~\cite{gu2024anomalygpt} and Myriad~\cite{li2023myriad}, which fine-tunes LMMs to the IAD domain. AnomalyGPT uses an image decoder to predict pixel-level anomaly maps and then applies a prompt learner to guide large language models, but its reliance on coarse-level image tokens and predicted masks can lead to loss of fine-grained details. Myriad leverages pretrained IAD models as vision experts to resample fine-grained features via Q-Former~\cite{li2023blip}, effectively transferring general-purpose visual perception into the IAD domain. 
However, their reliance on simulated anomaly data and limited vision-language alignment undermines their ability to capture fine-grained visual attributes and understand manufacturing processes, ultimately leading to suboptimal performance.
To bridge these gaps, our work, Triad, first enhances attribute perception by introducing the CoT-M and EG-RoI, allowing manufacturing-driven anomaly detection.
\section{Methods}

\subsection{Problem definition}

Manufacturing-driven industry anomaly detection aims to determine whether an industrial product exhibits anomalies by jointly analyzing visual features and detailed product-specific manufacturing process descriptions. Given an input image and a textual description of the manufacturing process, the task requires the model to output a binary decision~(normal or anomalous).

While standard IAD tasks (e.g., \cite{gu2024anomalygpt, li2023myriad}) primarily focus on visual cues, manufacturing-driven IAD leverages intricate manufacturing process details, such as multi-step production procedures, to support a deeper analysis. This paradigm demands accurate visual attribute recognition and sophisticated reasoning that accounts for the interplay between visual cues and contextual manufacturing information, allowing more precise detection of defects arising from complex process deviations.

\begin{figure*}
\centering
\includegraphics[width=0.95\linewidth]{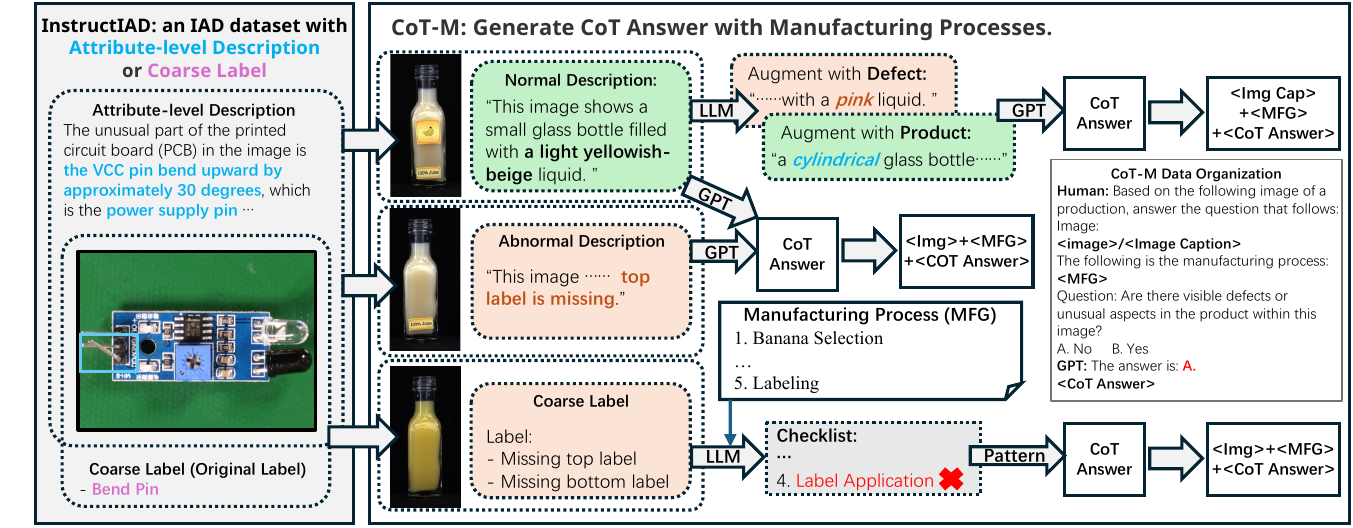}
\caption{Overview of the proposed CoT-M data organization pipeline.
Given normal product descriptions and manufacturing process information~(top row), the large language model expands product and defect types by editing product attributes, generating textual CoT data. When incorporating visual input, there are two scenarios: images with detailed textual descriptions (middle row) and images annotated only with coarse labels~(bottom row), collectively forming multimodal CoT data.}
\label{fig:data_generation}\vskip -0.2in
\end{figure*}

\subsection{Training Objectives}
\label{sec:train_obj}
Based on the problem defined before, our input, noted as $X$, is composed of vision part $X_V$ and language part $X_L$, with language annotation $Y$ having $n$ tokens $(y_1,y_2,...,y_n)$ from our training dataset, including InstructIAD and CoT-M.
$X_V$ is composed of query image $X_{img}$ and extra patches produced by EG-RoI. 
\begin{align}
X_V&=EG\mbox{-}RoI(X_{img},Expert(X_{img})).
\end{align}
Our fine-tuning procedure follows the common setting of Supervised Fine-Tuning~(SFT), including AdamW~\cite{adamw} optimizer and Cross Entropy loss function $\mathcal{L}_{SFT}$. 
\begin{align}
\label{eq: l_sft}
\mathcal{L}_{SFT}&=-\sum_{i=1}^{n}{\log \pi_{\theta}(y_i|X,y_1,y_2,...,y_{i-1})},
\end{align}
where $\theta$ represents the parameters of LMMs and $\pi_{\theta}(y|X)$ represents the probability of LMM generating the response $y$ given input $X$. 
The whole training procedure could be formulated as optimize $\mathcal{L}_{SFT}(LMM(X_V,X_L,\theta),Y)$.
Our fine-tuning involves all parameters of LMM and has only one stage with all tasks mixed from both InstructIAD and CoT-M. Technique details will be demonstrated in \cref{sec: experiments}.

\subsection{InstructIAD Dataset}
\label{sec:instructIAD}
We first collect an instruction-tuning dataset with human-annotated attribute-rich captions and coarse labels from existing IAD datasets, RealIAD~\cite{realiad}, VisA~\cite{visa}, and MVTec-LOCO AD~\cite{mvtecloco}, as shown in the left part of \cref{fig:data_generation}.
InstructIAD comprises 9,444 abnormal and 13,578 normal samples spanning 40 different products in RealIAD and VisA. From these, we manually annotate 2,098 samples (1,049 normal and 1,049 defect images) across all 40 classes, providing fine-grained, attribute-level captions describing products (\eg, color, shape, layout, material, texture) and defects (\eg, location, orientation, shape, color).
InstructIAD supports three key tasks:
i)~Anomaly detection – a binary (Yes/No) classification task for samples that only have normal or abnormal labels,
ii)~Attribute-level caption – similar to standard image caption but with a heightened focus on detailed product and defect attributes, and
iii)~Anomaly analysis – a two-part task where the model predicts whether an image is normal or abnormal, then provides an explanation grounded in the relevant visual attributes. Explanations are generated by LLaMA3~\cite{llama3} from the attribute-level descriptions.
Together, these tasks build a logical path from fine-grained visual attributes to anomaly detection.
Please refer to \cref{sec:task_details} in the supplementary material for details.

\subsection{Chain-of-Thought with Manufacturing}
\label{sec:CoT-M}
To achieve manufacturing-aware reasoning, we introduce CoT-M: a data organization strategy that infuses Chain-of-Thought~(CoT) reasoning with manufacturing processes. CoT-M is built upon InstructIAD and aims to enhance anomaly detection by joint reasoning between product attributes and manufacturing-related information.

CoT-M extends InstructIAD along two axes: (i)~product and defect diversity and (ii)~manufacturing awareness. 
The data-generation pipeline (\cref{fig:data_generation}) adapts to the information available for each sample through three complementary modes: (a)~Images with captions (\cref{fig:data_generation}, middle): The original attribute-level caption and its MFG are fed into GPT, which interleaves visual details with manufacturing steps to yield a coherent chain-of-thought (CoT). The outcome is an annotated triplet (image, manufacturing process, reasoning trajectory);
(b)~Images without captions (\cref{fig:data_generation}, bottom): For categories such as ``juice bottle'' in MVTec-LOCO AD, caption information is absent. We therefore invoke a checklist-style template that emulates a human inspector. Each coarse defect label is cross-referenced with an LLM-generated MFG checklist to pinpoint the faulty visual attribute; the completed checklist itself serves as the CoT explanation.
(c)~Text-only exemplars (\cref{fig:data_generation}, top):
Starting from a normal product caption in InstructIAD, we synthetically augment the product by altering attributes (\eg, colour, component type, quantity) and stochastically injecting defect descriptions. GPT then generates the reasoning trajectory conditioned on the augmented caption and its MFG, producing a purely textual CoT pair.
After generating the accompanying reasoning trajectories with the augmented descriptions and manufacturing processes, we manually filter out any hallucinated or erroneous samples. The final output, including an augmented caption, corresponding manufacturing processes, and generated reasoning process forms a new textual CoT answer pair.

\subsection{Expert-guided RoI module}
\begin{figure}
\centering
\includegraphics[width=1.0\linewidth]{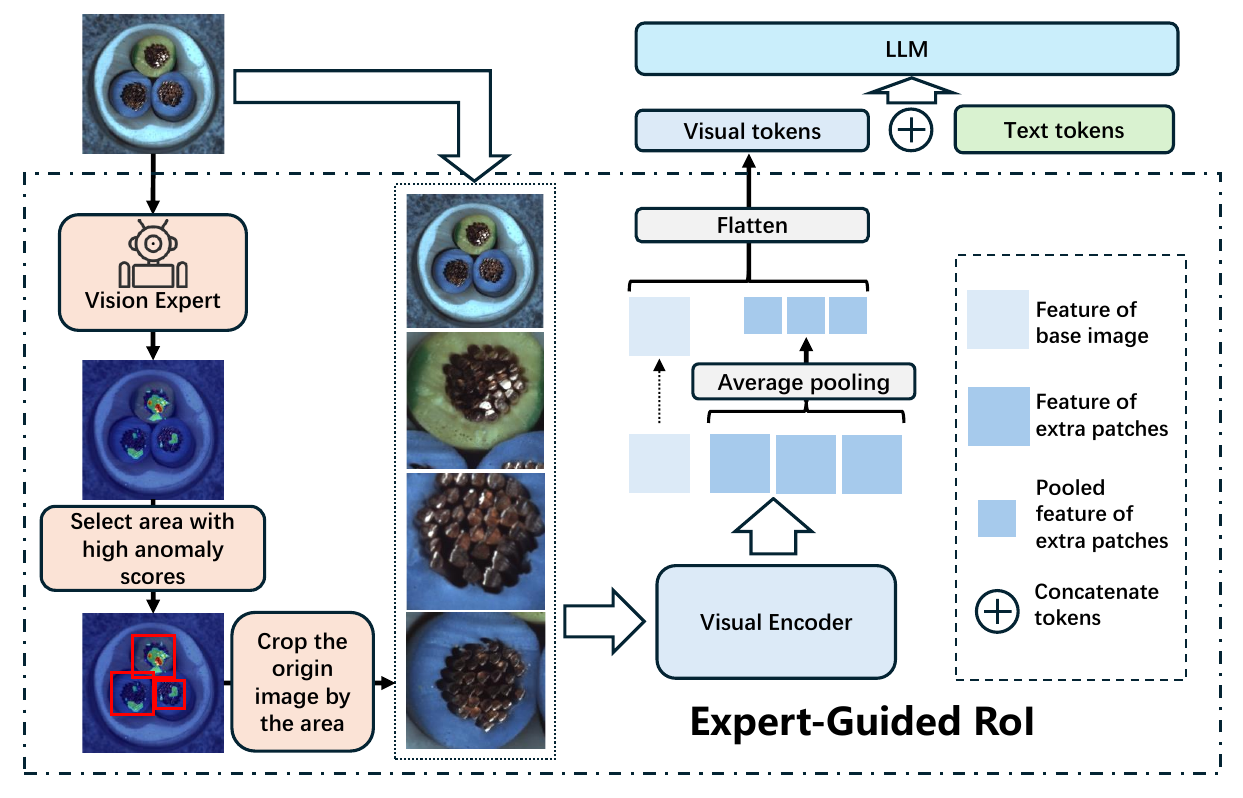}
\caption{The whole workflow of EG-RoI, with cropped potential defective area encoded as extra patches, Triad could have more detailed visual information to inspect.}
\label{fig: egroi}\vskip -0.2in
\end{figure}
In LMMs, the image is encoded into tokens by a visual encoder such as CLIP~\cite{clip} and SigLIP~\cite{siglip}, thus the encoder resizes the resolution of the input images to a fixed size. To offset the effect of limited resolution, LLaVA-Next~\cite{llavanext} has proposed a technique called AnyRes by entering extra large patches of the original image. This technique improves performance and has been further developed in LLaVA-OneVision~\cite{llavaov}. However, dividing the whole image into patches as extra patches is memory-consuming and redundant for IAD, as most patches of a defective image do not include defects. 
Accordingly, we propose the EG‑RoI module (see \cref{fig: egroi}), which improves the recognition of product attributes and defects. During training, ground‑truth defect regions together with randomly sampled normal regions are supplied to the module. During the evaluation, we can leverage established zero‑shot IAD vision experts—MuSc~\cite{li2024musc}, AnomalyCLIP~\cite{zhou2023anomalyclip}, and April‑GAN~\cite{chen2023april}—to supply informative suspicious regions to EG‑RoI. 
%
Anomaly maps from the vision experts are normalized and binarized; pixels scoring $>0.9$ define suspicious regions. Each region is enclosed in a fixed‑size box cropped from the image to preserve resolution. To comply with the language model’s context‑length limit, heavily overlapping boxes are merged, and the total number of regions is capped at four.
The resized original image forms the base view, while cropped regions serve as auxiliary patches. 
Both the base image and the patches are processed by the image encoder; patch tokens are average‑pooled to save memory, flattened, and concatenated with the base‑image tokens. The combined visual tokens are then projected into the textual embedding space and passed as a single sequence to the language model.
By supplying these high‑resolution local views, EG‑RoI allows the LMM to conduct a more detailed MFG-driven inspection.

\section{Experiments}
\label{sec: experiments}
\begin{table}
  \centering
  \small
  \renewcommand\arraystretch{1.2}
  \caption{Zero-shot anomaly detection performance with \textbf{manufacturing  process}~(+MFG Proc.) on MVTec-AD~\cite{bergmann2019mvtec} and WFDD~\cite{wfdd} datasets. The best results are in \textbf{bold} while the second best are \underline{underlined}. }
  \vspace{-2mm}
  \resizebox{\linewidth}{!}{
  \begin{tabular}{@{}c|c|cc|cc@{}}
  \toprule
  \multirow{2}{*}{Model} & \multirow{2}{*}{Params} & \multicolumn{2}{c|}{MVTec-AD} & \multicolumn{2}{c}{WFDD} \\
  & & 0-shot  & + MFG Proc. & 0-shot & + MFG Proc. \\ \midrule
  GPT-4o~\cite{gpt4o}  & - & 82.2\% & 67.9\%~\stdvd{14.3\%} & 78.5\% & 77.3\%~\stdvd{1.2\%}  \\ \midrule
  Qwen2-VL~\cite{qwen2vl}  & 2B & 77.0\% & 46.7\%\stdvd{30.3\%}  
                            & 70.6\% & 45.2\%\stdvd{25.4\%} \\ 
  LLava-1.6~\cite{llavanext} & 7B & 76.9\% & 75.9\%~\stdvd{1.0\%} 
                            & 63.8\% & 64.0\%~\stdvd{0.2\%} \\
  MiniCPM-V~\cite{minicpm_v}   & 8B   & 62.3\% & 51.6\%\stdvd{10.7\%}  & 70.3\% & 52.1\%\stdvd{18.2\%}  \\ 
  LLaVA-OneVision-si~\cite{llavaov} & 7B & 77.7\% & 60.6\%\stdvd{17.1\%} 
                            & 65.2\% & 61.4\%~\stdvd{3.8\%}  \\ 
  LLaVA-OneVision-ov~\cite{llavaov}   & 7B   & \underline{91.0\%} & 80.8\%\stdvd{10.2\%} 
                            & 79.8\% & \underline{80.3\%~\stdvu{0.5\%}} \\
  Qwen2-VL~\cite{qwen2vl}  & 7B & 84.4\% & 61.1\%\stdvd{23.3\%} 
                            & 74.4\% & 61.4\%\stdvd{13.0\%} \\ 
  Qwen2-VL~\cite{qwen2vl}  & 72B & 87.1\% & 79.5\%~\stdvd{7.6\%} 
                            & \textbf{81.1\%} & 74.2\%~\stdvd{6.9\%} \\
  LLaVA-OneVision-ov~\cite{llavaov}  & 72B   & {87.3}\% & 75.5\%\stdvd{11.8\%} 
                            & 75.0\% & 74.6\%~\stdvd{0.4\%} \\\midrule
  Myriad~\cite{li2023myriad}      & 7B & 79.3\% & 81.5\%~\stdvu{2.5\%}
                            & 60.5\% & 61.7\%~\stdvu{1.2\%} \\ \midrule
  \cellcolor{cellpink} Triad-llava-1.6 & 
  \cellcolor{cellpink}7B & 
  \cellcolor{cellpink}85.0\% & 
  \cellcolor{cellpink}\underline{87.5\%~\stdvu{2.5\%}} & 
  \cellcolor{cellpink}67.3\% & 
  \cellcolor{cellpink}69.9\%~\stdvu{2.6\%} \\ 
  \cellcolor{cellpink} Triad-ov  &
  \cellcolor{cellpink}7B & 
  \cellcolor{cellpink}\textbf{91.2\%} & 
  \cellcolor{cellpink}\textbf{92.6\%~\stdvu{1.4\%}} & 
  \cellcolor{cellpink}\underline{80.2\%} & 
  \cellcolor{cellpink}\textbf{81.1\%~\stdvu{0.9\%}} \\ \bottomrule
  \end{tabular}}\vskip -0.1in

  \label{tab: zero-shot}\vskip -0.1in
\end{table}

Our experiments mainly focus on how manufacturing processes boost LMMs in IAD tasks through our method, including common results from general LMMs and LMM-based IAD methods and ablation experiments of different industrial contexts.

\noindent\textbf{Implementation Details}
Our method is implemented on both LLaVA-1.6 (the earlier version of LLaVA-NeXT~\cite{llavanext}) and LLaVA-OneVision-ov (the checkpoint after the ``one-vision stage'' of LLaVA-OneVision~\cite{llavaov}), referring to them as Triad-llava-1.6 and Triad-ov, respectively. We build our approach on the LLaVA architecture by integrating our expert-guided region-of-interest~(EG-RoI) tokenizer. Specifically, in the LLaVA-1.6 version, the original AnyRes module is replaced with the EG-RoI module. In contrast, for LLaVA-OneVision, we append the suspicious regions to the output of the AnyRes module to fully leverage its native anomaly detection capabilities.

To preserve the generalization of the base models, we supplement our IAD-related instruction data by sampling 12K pairs from the original fine-tuning datasets of both LLaVA-1.6 and LLaVA-OneVision. Moreover, since LLaVA-1.6 lacks inherent multi-image processing capabilities, we constructed a simple dual-image caption dataset using the COCO subset~\cite{coco} from the ShareGPT4V dataset~\cite{sharegpt4v}. This dataset is exclusively used in the 1-shot setting to provide basic multi-image support.
Given the distinct objectives of 0-/1-shot anomaly detection, we offer separate versions of Triad for each setting, with data organized by CoT-M with specific instructions. For the 1-shot version of Triad-llava-1.6, we integrate the zero-shot model with the one-shot model using a Confidence Voting Mechanism (see \cref{sec:cvm}) to mitigate its multi-image inability.

Both Triad-llava-1.6 and Triad-ov were trained on 4×A800 80G GPUs with a mega-batch size of 128. For Triad-llava-1.6, we set the per-device batch size to 8 with a 4-step gradient accumulation. In the case of Triad-ov, due to the context length increasing from 4096 to 32768 tokens, the per-device batch size was reduced to 1, and gradient accumulation was increased to 32 steps. All other settings follow those established for LLaVA-1.6 and LLaVA-OneVision. With these configurations, 0-shot fine-tuning requires approximately 3 hours, while 1-shot fine-tuning takes about 5 hours for Triad-llava-1.6 and roughly twice that for Triad-ov.

\noindent\textbf{Evaluation Details}
We evaluate Triad using images from MVTec-AD~\cite{bergmann2019mvtec}, WFDD~\cite{wfdd}, and PCB-Bank~\cite{pcb_bank} for quantity and quality results, complemented by product-specific manufacturing processes. Since the original datasets do not provide meta-information about their products, we employ ChatGPT4 to generate manufacturing processes based on each product’s name and a caption describing a normal (non-defective) item. In real-world industrial applications, these manufacturing process details would typically be supplied directly by the factory.

Our evaluation comprises 21 different products and a total of 3003 multiple-choice questions following recent multimodal benchmarks~\cite{mmbench,mmmu,seedbench}, spanning both object and texture categories. To broaden the complexity of anomaly detection scenarios, we incorporate WFDD and PCB-Bank, which introduce diverse texture-based and object-based defect types, respectively.
Similarly, we use multiple-choice accuracy as the metric.

For zero-shot evaluation, the instruction follows the same format as the anomaly detection task shown in \cref{fig:data_generation}.
For one-shot evaluation, general LMMs that support multi-image directly input the query and a normal reference image with a similar prompt with zero-shot, only the question is replaced with: 

\textit{The second image shows an acceptable product. Compared to the acceptable product, find out whether there are defects in the product in the first image.} 

LMM-based IAD method AnomalyGPT~\cite{gu2024anomalygpt} and Myriad~\cite{li2023myriad} are tested according to their own instructions. Because of their low instruction following ability, we retrieve the keyword \textit{yes} or \textit{no} for accuracy calculation. Detailed demonstration can be found in \cref{sec: extra_eval}.

\noindent\textbf{Baseline}
In this study, we perform extensive evaluations against several robust baselines, including AnomalyGPT~\cite{gu2024anomalygpt} and Myriad~\cite{li2023myriad}. 
General-purpose LMMs exhibit enhanced reasoning capabilities as a result of being fine-tuned on complex reasoning data spanning diverse domains, including mathematics, chart interpretation, and document analysis. For evaluation, we benchmark our approach against state-of-the-art general-purpose LMMs, including Qwen2-VL~\cite{qwen2vl}, MiniCPM-V~\cite{minicpm_v}, the LLaVA series~\cite{llavanext, llavaov}, and the closed-source GPT-4o~\cite{gpt4o}.

\subsection{Main results}
\label{sec:main_results}

\noindent\textbf{Quantity Results}
The zero-shot anomaly detection results are presented in \cref{tab: zero-shot}. Although state-of-the-art LMMs such as LLaVA-OneVision-ov-7B~\cite{llavaov} and Qwen2-VL-72B~\cite{qwen2vl} demonstrate strong zero-shot performance on MVTec-AD, WFDD, and PCB-Bank~(see \cref{sec:pcb_bank}), our Triad-ov-7B remains highly competitive. Notably, by integrating manufacturing processes, Triad-ov-7B surpasses LLaVA-OneVision-ov-7B by 1.6\%, highlighting the advantage of manufacturing-aware reasoning in IAD. In contrast, most general-purpose LMMs fail to integrate manufacturing information, resulting in a significant performance drop. Interestingly, we also observe that Myriad benefits from manufacturing processes, likely due to its built-in visual enhancement mechanisms.

To evaluate the adaptability of our approach, we provide two versions of Triad based on different LMM backbones, LLaVA-1.6-7B and LLaVA-OneVision-ov-7B. When starting with the relatively weak LLaVA-1.6-7B model, Triad yields substantial gains of 8.1\% (from 76.9\% to 85.0\% on MVTec-AD) and 3.5\% (from 63.8\% to 67.3\% on WFDD). Moreover, equipping Triad-llava-1.6-7B with manufacturing processes adds a further 2.5\% and 2.6\% improvement, respectively. Similarly, Triad-ov-7B outperforms the original LLaVA-OneVision-ov-7B and gains an additional 1.4\% and 0.9\% boost from manufacturing-aware reasoning on MVTec-AD and WFDD, respectively. These results confirm that our approach generalizes effectively across different model architectures and continues to enhance performance even at higher baseline accuracies (e.g., above 90\%).

\Cref{tab:one-shot} shows 1-shot results. Providing a single reference image generally degrades the performance of most general-purpose LMMs due to their limited instruction-following capabilities. Two exceptions, LLaVA-OneVision-ov-7B (91.5\% vs. 91.0\% in 0-shot) and Qwen2-VL-72B (92.0\% vs. 87.3\% in 0-shot), still cannot fully utilize manufacturing information for anomaly detection~(accuracy down by 5.7\% and 1.6\%, respectively). Meanwhile, Triad-llava-1.6-7B achieves comparable performance to both AnomalyGPT and Myriad under 1-shot conditions, and significantly outperforms them with manufacturing processes (88.4\% vs. 85.4\% for Myriad and 80.9\% for AnomalyGPT). 
Leveraging LLaVA-OneVision-ov-8B—a carefully tuned multi-image baseline—Triad-ov-8B achieves 92.9\% on MVTec-AD in the 1-shot setting, further improving to 94.1\% when incorporating manufacturing processes. These results suggest that even a strong base model can benefit from the Manufacturing processes. 
Overall, our findings demonstrate that Triad offers robust and flexible anomaly detection across both zero-shot and few-shot scenarios, while the manufacturing-driven IAD paradigm consistently delivers strong performance gains.

\begin{table}
  \centering
  \small
  \renewcommand\arraystretch{0.9}
  \caption{One-shot anomaly detection performance with \textbf{manufacturing process} on MVTec-AD~\cite{bergmann2019mvtec} The best results are in \textbf{bold} while the second best are \underline{underlined}.}
  \vspace{-3mm}
  \resizebox{0.48\textwidth}{!}{
  \begin{tabular}{@{}c|c|cc@{}}
  \toprule
  \multirow{2}{*}{Model} & \multirow{2}{*}{Params} & \multicolumn{2}{c}{MVTec-AD} \\
  & & 1-shot & + MFG Proc. \\ \midrule
  GPT-4o~\cite{gpt4o}  & - & 77.6\% & 72.5\%~\stdvd{5.1\%}  \\ \midrule
  Qwen2-VL~\cite{qwen2vl}  & 2B & 32.1\% & 30.7\%~\stdvd{1.4\%}  \\
  LLava-1.6~\cite{llavanext} & 7B & 72.3\% & 76.0\%~\stdvu{3.7\%}  \\ 
  LLaVA-OneVision-ov~\cite{llavaov}   & 7B   & 91.5\% & 85.2\%~\stdvd{5.7\%}  \\
  Qwen2-VL~\cite{qwen2vl}  & 7B & 81.7\% & 84.5\%~\stdvu{2.8\%} \\ 
  Qwen2-VL~\cite{qwen2vl}  & 72B & \underline{92.0\%} & \underline{90.4\%~\stdvd{1.6\%}}  \\ \midrule
  AnomalyGPT\footnotemark~\cite{gu2024anomalygpt} & 7B & 86.1\% & 80.9\%~\stdvd{5.2\%} \\
  Myriad~\cite{li2023myriad}& 7B & 87.4\% & 85.4\%~\stdvd{2.0\%}  \\ \midrule
  Triad-llava-1.6           & 7B & 87.7\% & 88.4\%~\stdvu{0.7\%} \\ 
  Triad-ov  & 7B & \textbf{92.9\%} & \textbf{94.1\%~\stdvu{1.2\%}} \\  \bottomrule
  \end{tabular}}\vskip -0.25in

  \label{tab:one-shot}
\end{table}
\footnotetext{AnomalyGPT adds hints mentioning defect types for each product in MVTec-AD test set.}

\begin{figure*}
  \centering
  \includegraphics[width=\linewidth]{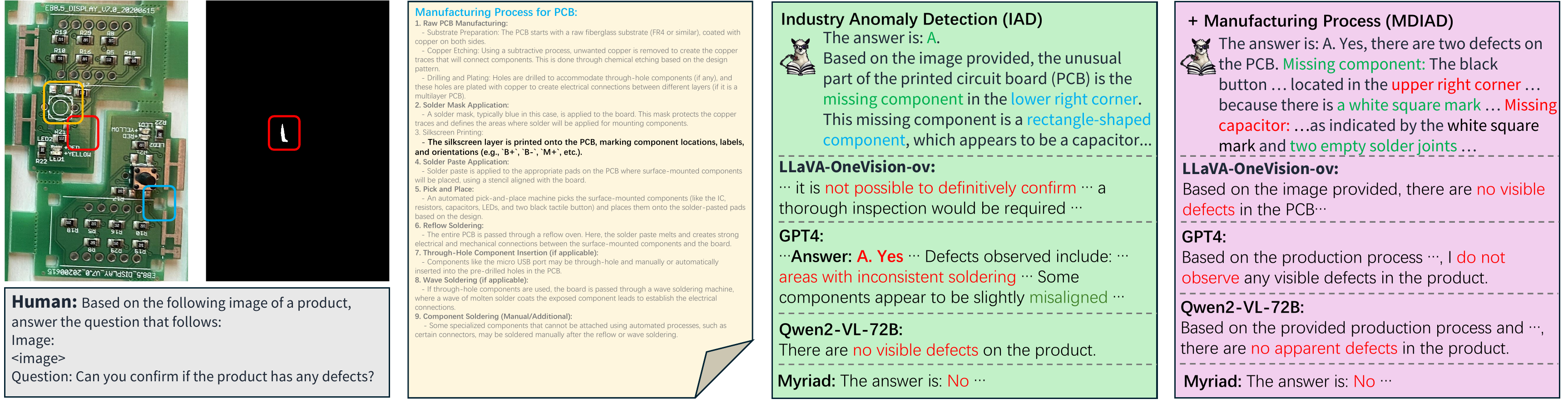}
  \caption{Comparison with and without manufacturing processes on a PCB7 example from PCB-Bank~\cite{pcb_bank}. The responses from state-of-the-art LMMs are provided. Expert's detection is indicated by a red bounding box, while the ground truth is shown in yellow. An additional component identified by Triad is highlighted with blue bounding boxes. Best viewed in color.}
  \label{fig:qualitative_2}
\end{figure*}

\begin{figure*}
\centering
\begin{minipage}{0.51\linewidth}
\begin{subfigure}{\linewidth}
    \centering
    \includegraphics[width=\linewidth]{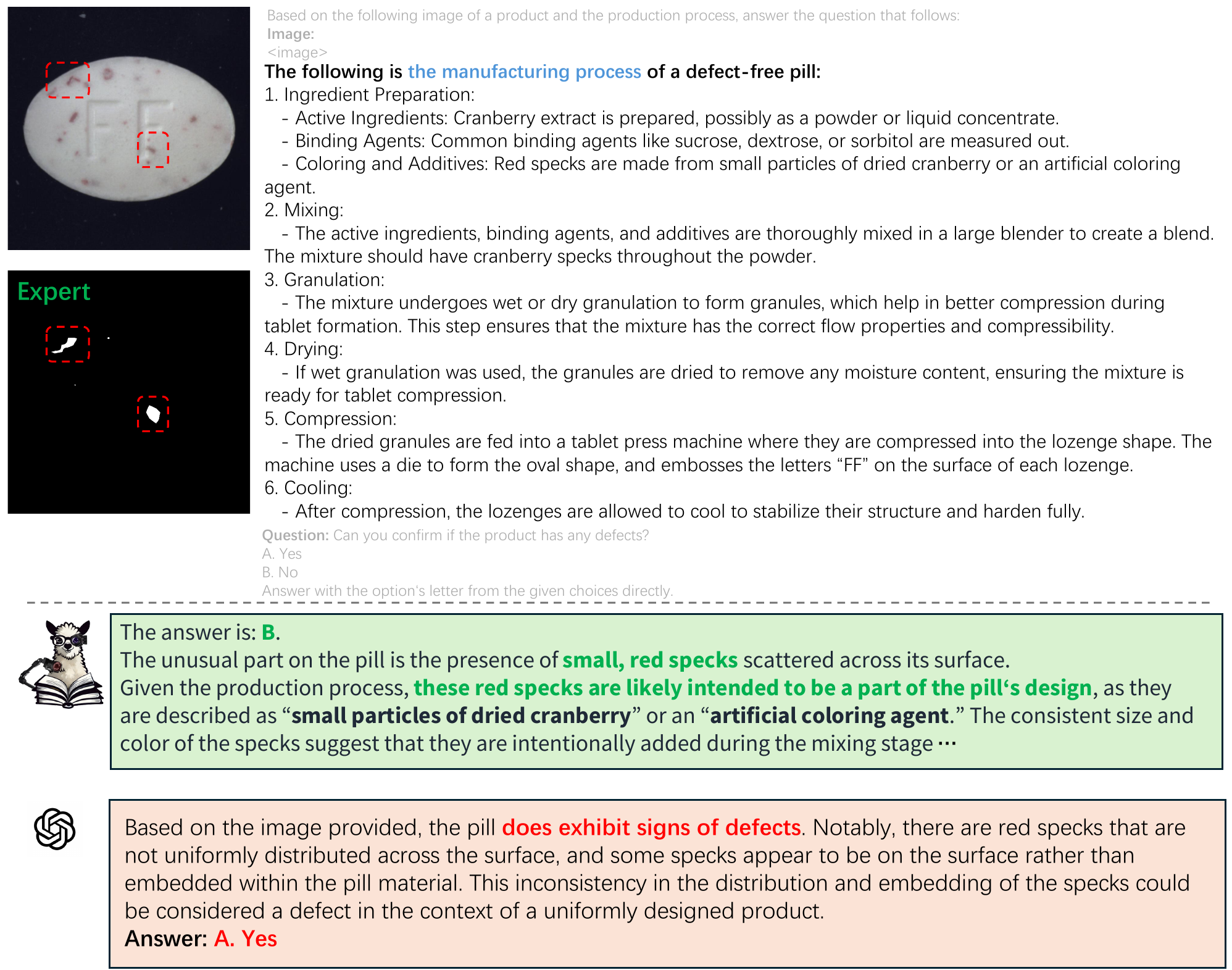}
    \caption{}
    \label{fig:qualitative_1_a}
\end{subfigure}
\end{minipage}
\begin{minipage}{0.47\linewidth}
\begin{subfigure}{0.49\linewidth}
    \centering
    \includegraphics[width=\linewidth]{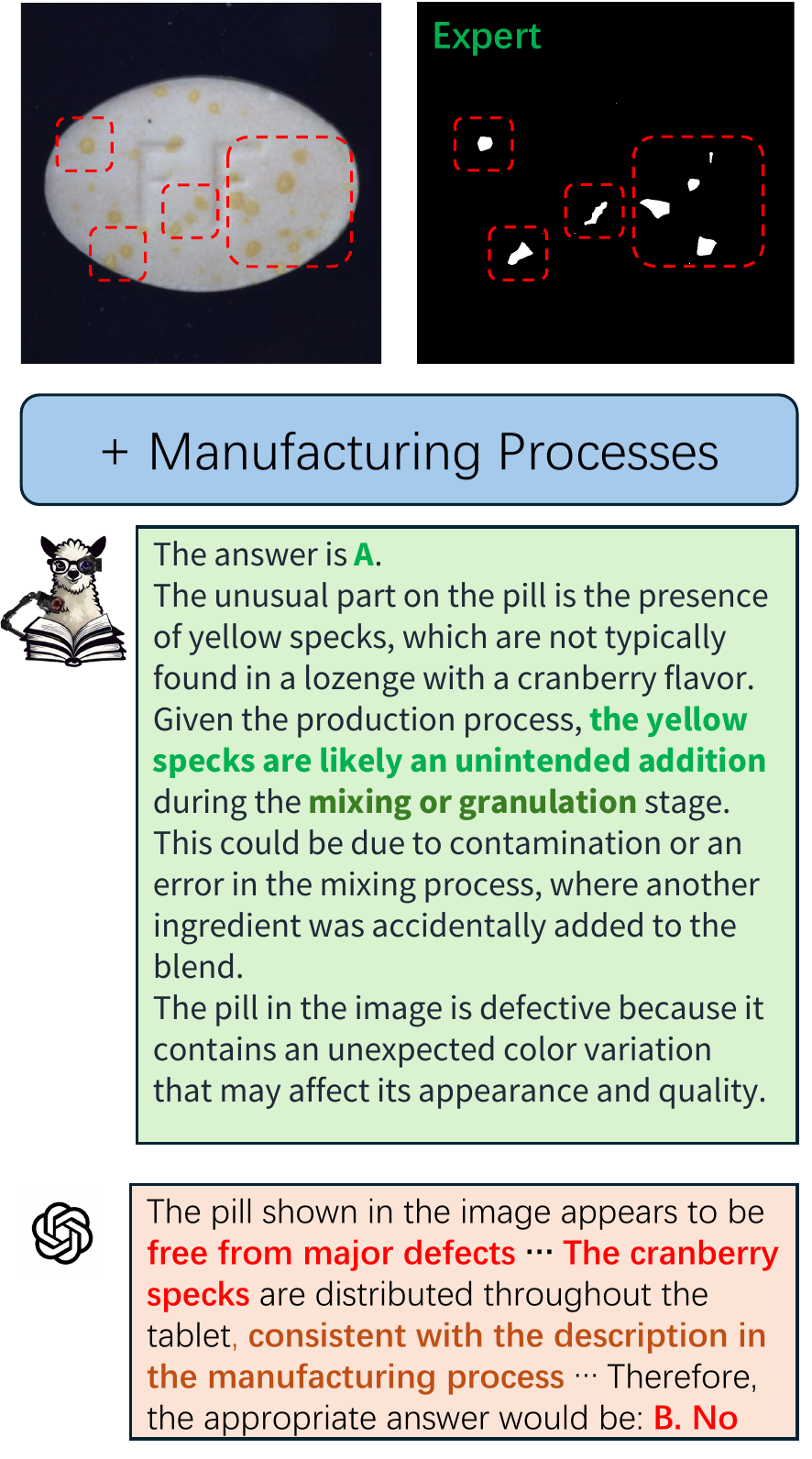}
    \caption{}
    \label{fig:qualitative_1_b}
\end{subfigure}
\begin{subfigure}{0.47\linewidth}
    \centering
    \includegraphics[width=\linewidth]{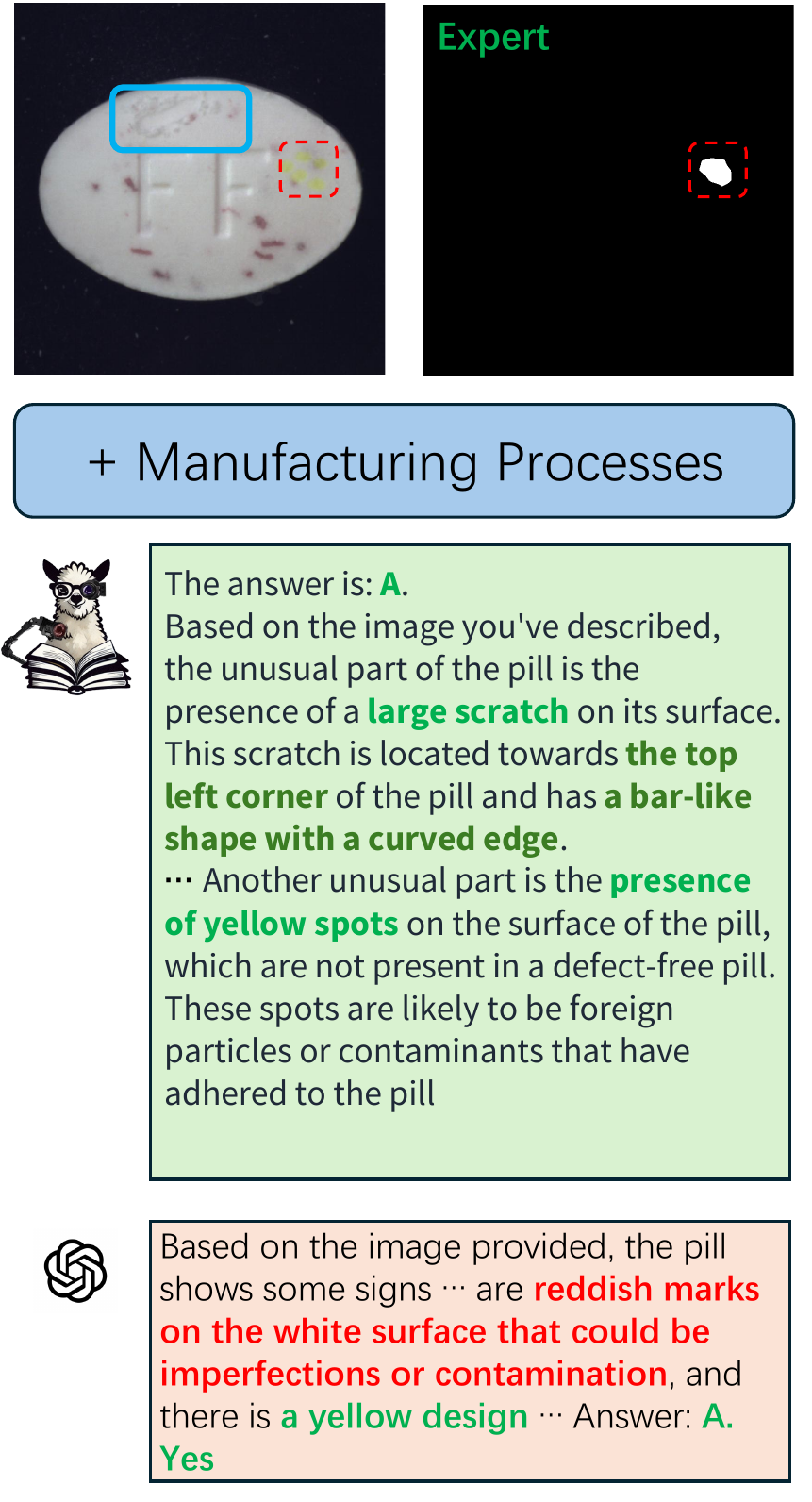}
    \caption{}
    \label{fig:qualitative_1_c}
\end{subfigure}
\end{minipage}
\vskip -0.1in
\caption{Qualitative evaluation of Triad-llava-ov-7B on MVTec-AD~\cite{bergmann2019mvtec}. MuSc~\cite{li2024musc} is used as vision expert. The related manufacturing process is shown in (a). Three representative examples from the ``Pill'' class are shown: \textbf{(a)}~a normal pill exhibits red speckles on its surface which are a natural outcome of the manufacturing process and could be easily misjudged as defects. \textbf{(b)}~Pills with unintended additive variations indicate a defect. \textbf{(c)} a case where Triad correctly predicts a defect despite errors in the expert outputs.}
\label{fig:qualitative_1}\vskip -0.2in
\end{figure*}

\noindent\textbf{Qualitative results.}
\Cref{fig:qualitative_2} compares IAD with and without manufacturing processes. In the standard IAD setting, Triad and GPT-4o both identify a board-level defect. However, Myriad incorrectly flags a different (normal) component, and GPT-4o infers inconsistent soldering without a clear visual cue. When manufacturing process information is included, Triad accurately detects the missing button by referring to the white square silkscreen mark, which indicates the location of the button. This demonstrates Triad’s manufacturing-aware reasoning capacity, even when handling previously unseen manufacturing descriptions.

\Cref{fig:qualitative_1} presents three representative cases. In the first example (\cref{fig:qualitative_1_a}), a surface discoloration might typically be viewed as a defect (as GPT-4o incorrectly predicts); however, Triad recognizes that the discoloration corresponds to the cranberry additive described in the manufacturing process, leading to a correct classification. Thanks to the EG-RoI module’s ability to capture fine-grained visual cues and the data organization via CoT-M, Triad confidently aligns the observed appearance with the relevant manufacturing steps.
In the second example (\cref{fig:qualitative_1_b}), a pill containing “yellow speckles” is mistakenly introduced into a production pipeline that should only produce pills with “red speckles.” Triad correctly identifies the inconsistency between the visual observation and the manufacturing procedures. GPT-4o, though aware that red speckles are the standard, fails to accurately match the color of the speckles in question. This underscores the necessity of precise attribute recognition for reasoning.
Finally, the third example (\cref{fig:qualitative_1_c}) highlights Triad’s resilience to vision expert errors, where it fails to detect a scratch on the surface. Triad still identifies the scratch independently and also notes a yellow discoloration, demonstrating its capacity to rectify external region proposal mistakes by integrating visual analysis with contextual defect reasoning.

Overall, these observations illustrate Triad’s ability to integrate manufacturing knowledge with detailed visual attributes, thereby achieving more accurate and context-sensitive anomaly detection.

\subsection{Ablations}
\label{sec:ablations}

\noindent\textbf{Effect of InstructIAD.}
InstructIAD is placed as an instruction-tuning dataset for IAD, which aims to align the visual features and textual descriptions, leading to a performance gain compared with the base model~(LLaVA-1.6-mistral-7B). Notably, as shown in \cref{tab: ablation}, the performance with manufacturing processes also increases, empirically demonstrating the intrinsic relationship between attribute-level recognition and manufacturing-aware defect analysis. 

\noindent\textbf{Effect of CoT-M.}
As evidenced by \cref{tab: ablation}, comparative analysis of row 2 vs. 3 and row 4 vs. 5 demonstrates the efficacy of CoT-M. Utilizing CoT-M-generated data enhances manufacturing process detection accuracy by 2.1\% and 3.6\% respectively, while maintaining comparable/superior performance in standard detection scenarios compared to non-CoT-M approaches. This finding substantiates that CoT-M empowers Triad to effectively leverage manufacturing process information for improved anomaly detection.

\noindent\textbf{Effect of EG-RoI.}
Comparative analysis in \cref{tab: ablation} (row 3 vs. 5) confirms EG-RoI's consistent performance gains across both manufacturing-aware and conventional detection scenarios.
We further evaluate vision expert effectiveness through size-based defect categorization on MVTec-AD, adopting MS-COCO's partitioning protocol~\cite{coco}: small ($<0.01$ image area), medium ($0.01\sim0.1$), and large ($>0.1$). As shown in \cref{tab: ablation_roi}, EG-RoI significantly outperforms LLaVA-1.6's AnyRes module on small/medium defects. The baseline exhibits severe classification bias with 100\% defect detection rate but 16.1\% false positives on normal samples.
\Cref{tab: ablation_expert} reveals two critical findings: (1)~Anomaly map quality directly impacts detection accuracy, validating our expert integration strategy; (2) Triad achieves competitive performance even without annotated regions~(AnyRes mode), surpassing conventional methods. Notably, EG-RoI maintains manufacturing-aware improvements despite random box noise injection~(Null mode).

\noindent\textbf{Ablations on the different manufacturing process.}
For systematic evaluation, we collect three distinct manufacturing process variants for MVTec AD~\cite{bergmann2019mvtec} through multi-source construction: (1) Internet: Domain-specific processes collected from the internet for all 15 product categories; (2) LLM Generation: Automated process synthesis using general-purpose LLMs (LLaMA-3~\cite{llama3} and GPT-4~\cite{achiam2023gpt4}). As evidenced in \cref{tab: pp_ablation}, Triad demonstrates consistent performance gains across all process variants (0.7\%-2.5\% accuracy improved), empirically validating its dual capability in manufacturing process comprehension and generalization to unseen industrial workflows. Extended process examples are provided in \cref{sec: extra_eval}.

\begin{table}
  \centering
  \small
  \renewcommand\arraystretch{1.0}
  \caption{Ablations on fine-tuning with different components: InstructIAD, CoT-M, and EG-RoI tokenizer. The base model is LLaVA-1.6-mistral-7B~\cite{llavanext} }
  \vspace{-2mm}
  \resizebox{0.48\textwidth}{!}{
  \begin{tabular}{@{}ccc|ccccc@{}}
  \toprule
  InstructIAD & CoT-M & EG-RoI & 0-shot & + MFG Proc. \\ \midrule
  \XSolidBrush & \XSolidBrush & \XSolidBrush &  76.9\% &  75.9\%~\stdvd{1.0\%}  \\ 
  \CheckmarkBold & \XSolidBrush & \XSolidBrush &  78.8\% & 80.4\%~\stdvu{1.6\%} \\ 
  \CheckmarkBold & \CheckmarkBold & \XSolidBrush &  79.8\% & 82.5\%~\stdvu{2.7\%} \\ 
  \CheckmarkBold & \XSolidBrush & \CheckmarkBold &  85.4\% & 83.9\%~\stdvd{1.5\%}  \\ 
  \CheckmarkBold & \CheckmarkBold & \CheckmarkBold & 85.0\% & 87.5\%~\stdvu{2.5\%} \\ \bottomrule
  \end{tabular}} \vskip -0.10in

  \label{tab: ablation} 
\end{table}

\begin{table}
  \centering
  \small
  \renewcommand\arraystretch{1.2}
  \caption{Ablation on the robustness of the manufacturing process on MVTec-AD~\cite{bergmann2019mvtec}. Three types of the manufacturing processes of 15 products on MVTec-AD are collected from the Internet~(Web), generated by llama3~\cite{llama3}~(LLM), and generated by GPT4~\cite{achiam2023gpt4}~(GPT).}
  \vspace{-2mm}
  \resizebox{0.48\textwidth}{!}{
  \begin{tabular}{@{}c|c|cccc@{}}
  \toprule
  Model & Params & 0-shot & + MFG Proc.~(Web) & + MFG Proc.~(LLM) & + MFG Proc.~(GPT) \\ \midrule
  Qwen2-VL~\cite{qwen2vl}  & 2B & 77.0\% & 51.7\%\stdvd{25.3\%} & 46.7\%\stdvd{30.3\%} & 46.8\%\stdvd{30.2\%}  \\ 
  LLava-1.6~\cite{llavanext} & 7B & 76.9\% & 77.6\%~\stdvu{0.7\%} & 75.9\%~\stdvd{1.0\%} & 77.3\%~\stdvu{0.4\%}  \\
  LLaVA-OneVision-si~\cite{llavaov} & 7B & 77.7\% & 69.4\%~\stdvd{8.3\%} & 60.6\%\stdvd{17.1\%} & 67.2\%\stdvd{10.5\%}  \\ 
  LLaVA-OneVision-ov~\cite{llavaov}   & 7B   & 91.0\% & 88.8\%~\stdvd{2.2\%} & 80.8\%\stdvd{10.2\%} & 87.1\%~\stdvd{3.9\%}  \\
  Qwen2-VL~\cite{qwen2vl}  & 7B & 84.4\% & 70.9\%\stdvd{13.5\%} & 61.1\%\stdvd{23.3\%} & 67.1\%\stdvd{17.3\%}  \\ 
  Qwen2-VL~\cite{qwen2vl}  & 72B & 87.1\% & 85.2\%~\stdvd{1.9\%} & 79.5\%~\stdvd{7.6\%} & 82.6\%~\stdvd{4.5\%} \\ \midrule
  Myriad~\cite{li2023myriad}      & 7B & 79.3\% & 78.5\%~\stdvd{0.8\%} & 81.5\%~\stdvu{2.2\%} & 81.5\%~\stdvu{2.2\%}  \\ \midrule
  Triad-llava-1.6           & 7B & 85.0\% & 85.7\%~\stdvu{0.7\%} & 87.5\%~\stdvu{2.5\%} & 86.4\%~\stdvu{1.4\%}  \\ 
  Triad-ov  & 7B & 91.2\% & 91.8\%~\stdvu{0.6\%} & 92.6\%~\stdvu{1.4\%} & 92.2\%~\stdvu{1.0\%}  \\ 
  \bottomrule
  \end{tabular}}\vskip -0.10in

  \label{tab: pp_ablation}  
\end{table}
\iftrue
\begin{table}
  \centering
  \small
  \renewcommand\arraystretch{1.0}
  \caption{Comparison between different vision experts with the proposed EG-RoI. The model is based on LLaVA-1.6-mistral-7B~\cite{llavanext}. The quality of anomaly maps by experts on zero-shot IAD tasks is measured by Expert P-AUROC.}
  \vspace{-2mm}
  \resizebox{0.48\textwidth}{!}{
  \begin{tabular}{@{}c|c|cccc@{}}
  \toprule
  Vision Expert & Expert P-AUROC & base & + MFG Proc. \\ \midrule
  Null  & - &  83.3\% & 83.4\%~\stdvu{0.1\%}  \\ 
  AnyRes & - &  84.0\% & 85.1\%~\stdvu{1.1\%}  \\ 
  April-GAN~\cite{chen2023april}  & 87.6\% &  83.0\% & 85.0\%~\stdvu{2.0\%}\\
  AnomalyClip~\cite{zhou2023anomalyclip} & 91.1\% &  84.6\% & 86.1\%~\stdvu{1.5\%} \\ 
  MuSc~\cite{li2024musc}  & 97.3\% &  85.0\% & 87.5\%~\stdvu{2.5\%} \\ 
  \bottomrule
  \end{tabular}}\vskip -0.1in

  \label{tab: ablation_expert}
\end{table}
\fi

\begin{table}
  \centering
  \small
  \renewcommand\arraystretch{1.0}
  \caption{Ablation on how Expert-Guided RoI module affects the performance. Defects are divided into small, medium, and large according to their size. The model is based on LLaVA-1.6-mistral-7B. All results are tested without context.}
  \vspace{-2mm}
  \resizebox{0.48\textwidth}{!}{
  \begin{tabular}{@{}c|cccc|c@{}}
  \toprule
  Module & small defects & medium defects & large defects & normal & Accuracy \\ \midrule
  baseline (llava-1.6)~\cite{llavanext} &  100.0\% & 100.0\% & 100.0\% & 16.1\% & 76.9\% \\
  AnyRes (Finetune)~\cite{llavanext}  &  90.9\% & 81.0\% & 65.7\% & 82.9\% & 79.8\%\\
  EG-RoI (Finetune) &  95.5\% & 94.1\% & 81.8\% & 72.8\% & 85.0\% \\ \bottomrule
  \end{tabular}}\vskip -0.2in

  \label{tab: ablation_roi}
\end{table}

\section{Conclusion}
In this paper, we introduced \textbf{Triad}, a novel large multimodal model (LMM) tailored for industrial anomaly detection. 
By integrating an expert-guided region-of-interest tokenizer (EG-RoI) to highlight suspicious regions identified by existing IAD methods, Triad improves its ability to pinpoint subtle defects in complex industrial scenarios. 
In addition, we proposed a manufacturing-driven IAD paradigm that embeds causal knowledge of manufacturing processes into the model’s reasoning for anomaly detection. Specifically, we contribute an instruction-tuning dataset, \textit{InstructIAD}, and a Chain-of-Thought with manufacturing strategy, CoT-M, enabling Triad to reason about defect formation in relation to manufacturing steps. 
Experimental results on standard IAD benchmarks demonstrate that Triad achieves superior performance in 0-/1-shot settings compared to both general-purpose and domain-specific LMMs. Extensive evaluations reveal Triad's novel ability to leverage manufacturing processes to achieve improved anomaly detection. Qualitative results show Triad's significant reasoning ability based on attribute recognition and manufacturing comprehension. These findings confirm the significance of combining expert-guided region-of-interest tokenizer with manufacturing-aware reasoning for robust and interpretable anomaly detection.

We publicly release InstructIAD's dataset and CoT-M data organization to facilitate future research, bridging the critical gap between general-purpose LMMs and domain-specific industrial inspection needs. We believe this work lays a starting point for modern quality control systems that synergize human knowledge with multimodal AI reasoning.

\section*{Acknowledgement}
This work was supported by the National Key R\&D Program of China under Grant No. 2023YFA1008500.
{
    \small
    \bibliographystyle{ieeenat_fullname}
    \bibliography{main}
}

\clearpage
\setcounter{page}{1}
\setcounter{section}{0}
\renewcommand{\thesection}{\Alph{section}}
\setcounter{table}{0}
\renewcommand{\thetable}{A\arabic{table}}
\setcounter{figure}{0}
\renewcommand{\thefigure}{A\arabic{figure}}
\maketitlesupplementary
\def\tabularxcolumn#1{m{#1}}

\section{Extra Results}

\subsection{Results on PCB-Bank}
\label{sec:pcb_bank}
We also test results on PCB-Bank~\cite{pcb_bank}, which includes a bunch of PCB images in a real industrial scene. Because part of the images in PCB-Bank come from VISA~\cite{visa} and RealIAD~\cite{realiad}, which are overlapped with our training data, we use the rest subset of PCB-Bank to evaluate for maintaining the zero-shot setting. Detecting defects on PCB is complicated and needs a lot of experience and knowledge in electronics, which is lacking in the general LMMs' knowledge base. However, Triad still outperforms due to its ability to use extra knowledge from the detailed manufacturing processes of PCB products.
\begin{table}[H]
  \centering
  \small
  \renewcommand\arraystretch{1.2}
  \caption{Zero-shot Anomaly Detection Performance with \textbf{manufacturing  process} on PCB-BANK~\cite{pcb_bank}. 
  }
  \vspace{-2mm}
  \resizebox{\linewidth}{!}{
  \begin{tabular}{@{}c|c|cc@{}}
  \toprule
  \multirow{2}{*}{Model} & \multirow{2}{*}{Params} & \multicolumn{2}{c}{PCB-Bank~(Subset)} \\ 
  & & 0-shot  & + MFG Proc. \\ 
  \midrule
  Qwen2-VL~\cite{qwen2vl}  & 2B & 49.9\% & 61.0\%~\stdvu{11.1\%} \\ 
  LLava-1.6~\cite{llavanext} & 7B & 38.9\% & 59.9\%~\stdvu{21.0\%} \\ 
  MiniCPM-V~\cite{minicpm_v}  & 8B & 61.2\% & 61.0\%~\stdvd{0.2\%} \\ 
  LLaVA-OneVision-si~\cite{llavaov} & 8B & 60.8\% & 61.0\%~\stdvu{0.2\%} \\
  LLaVA-OneVision-ov~\cite{llavaov}   & 8B & 59.6\% & 60.9\%~\stdvu{1.1\%} \\ 
  Qwen2-VL~\cite{qwen2vl}  & 7B & 59.5\% & 61.1\%~\stdvu{1.6\%} \\ 
  Qwen2-VL~\cite{qwen2vl}  & 72B & 59.6\% & 61.0\%~\stdvu{1.4\%} \\ 
  LLaVA-OneVision-ov~\cite{llavaov}  & 72B & 60.9\% & 61.4\%~\stdvu{0.5\%} \\ \midrule
  Myriad~\cite{li2023myriad}      & 7B & 61.8\% & 61.8\%~\stdvu{0.0\%} \\ \midrule
  Triad-llava-1.6 & 7B & 63.6\% & 64.8\%~\stdvu{1.2\%} \\   
  Triad-ov & 7B & 62.8\% & 63.8\%~\stdvu{1.0\%} \\   
  \bottomrule
  \end{tabular}}
  \label{tab:zero-shot}
\end{table}

\subsection{Results on MMAD}

We report results on MMAD~(\textbf{categories for training are filtered out}) in \cref{tab:mmad}. Triad still achieves the best performance on MMAD and outperforms LLaVA-OV by 9\% in both 0/1-shot. Note that the references for 1-shot evaluation on MMAD~(especially for GoodsAD) are not actually the same product as the query image. Therefore, all models do not improve much under 1-shot setting. 

\begin{table}
  \centering
  \renewcommand\arraystretch{0.6}
  \caption{0/1-shot anomaly detection performance on MMAD. }\vspace{-2mm}
  \resizebox{\linewidth}{!}{
  \begin{tabular}{c|ccc|c}
  \toprule
  Model & Qwen2-VL-7B & Qwen2.5-VL-7B & LLaVA-OV-ov & \cellcolor{cellpink} Triad-ov \\ \midrule
  0-shot & {56.25\%~~} & 50.14\%~~ & 60.74\%~~ & \cellcolor{cellpink} \textbf{70.22\%~~} \\ 
  + MFG Proc. & 49.50\%{\color{red}$\downarrow$} & 45.58\%{\color{red}$\downarrow$} & 59.60\%{\color{red}$\downarrow$} & \cellcolor{cellpink} \textbf{71.02\%}{\color{ForestGreen}$\uparrow$} \\ 
  \midrule
  1-shot & 57.78\%~~ & 42.71\%~~ & 60.80\%~~ & \cellcolor{cellpink} \textbf{70.92\%~~} \\ \bottomrule
  \end{tabular}}
  \label{tab:mmad}
  \vspace{-0.2 in}
\end{table}

\subsection{Extensive qualitative results}

In this section, more qualitative examples are provided to further illustrate how the manufacturing process helps anomaly detection and how Triad interacts with the changing manufacturing processes. %

\begin{figure}
\centering
\includegraphics[width=0.9\linewidth]{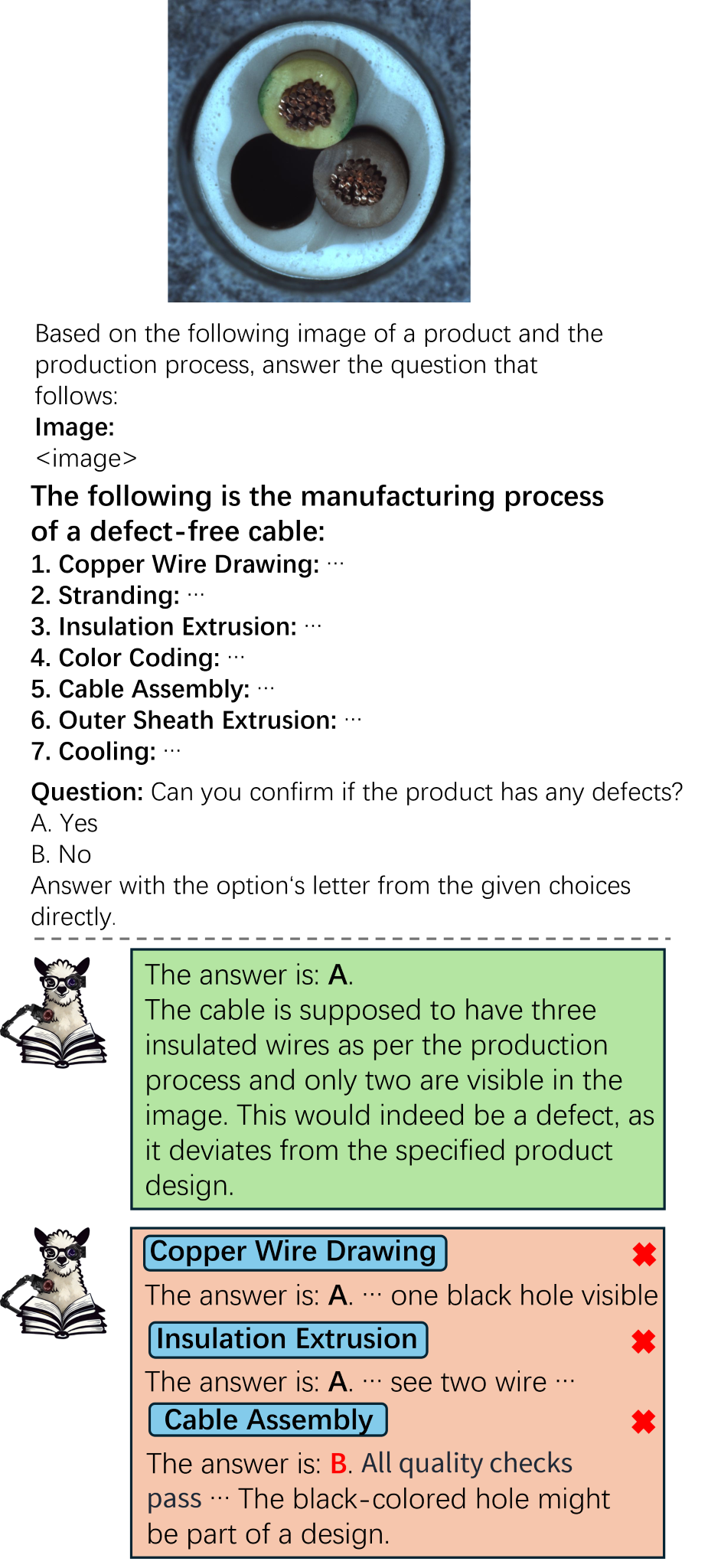}
\caption{Ablation on steering the manufacturing process of the cable. The model is based on llava-1.6. }
\label{fig:qualitative_3}
\end{figure}

In \cref{fig:qualitative_3}, we provide an example of the cable in MVTec-AD. The full manufacturing process is shown in \cref{tab:cable_mfg}. Four responses from Triad are presented: one includes the full process, and three omit certain information. The steps involving copper wire drawing and insulation extrusion do not specify the number of wires; therefore, their omission does not affect Triad's judgment regarding the missing wire cable. However, when the cable assembly step indicates there are three cables and explains how they are assembled, Triad adjusts its detection and considers the visible black-colored hole~(the absence of a wire) as part of the design.

\begin{figure*}
  \centering
    \centering
    \includegraphics{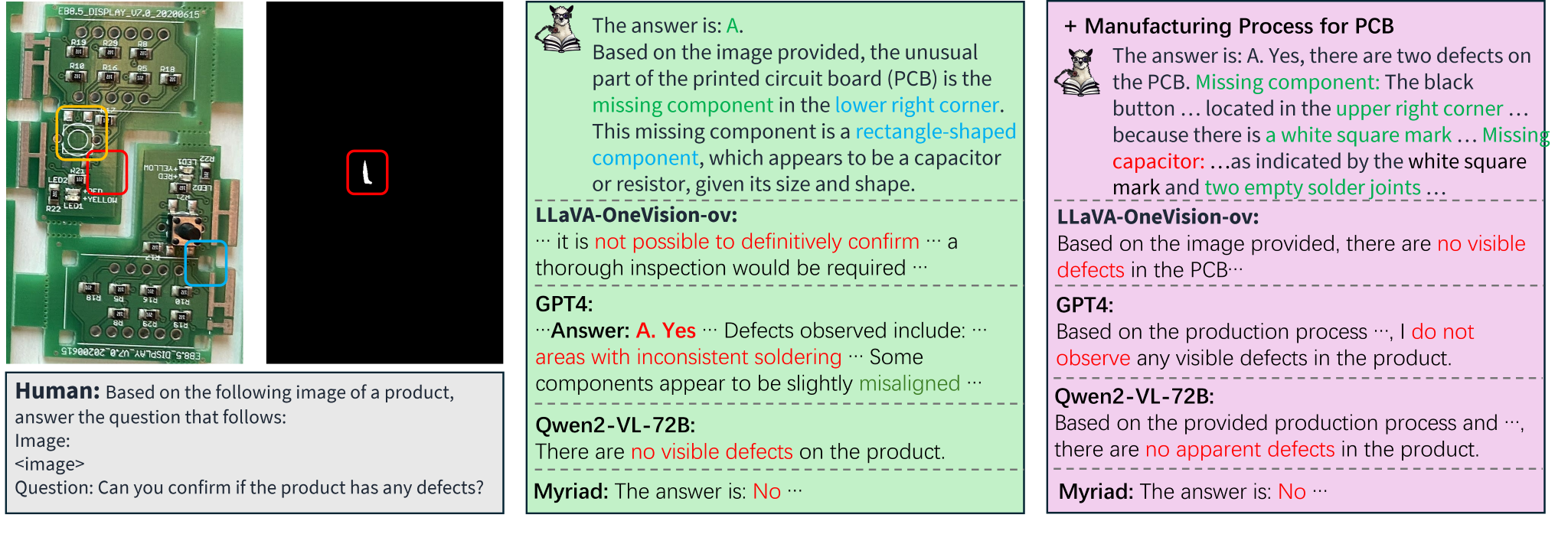}
  \caption{Qualitative results on another examples of PCB7 in PCB-Bank~\cite{pcb_bank}. The comparison between SOTA LMMs and one of the LMM-based IAD methods is presented. Our chosen vision expert~(\cite{li2024musc}) makes mistakes. The bounding box used by Triad is plotted in red, while the ground truth is plotted in yellow. Another component noticed by Triad is highlighted by blue bounding boxes. Best view in \textbf{color}.}
  \label{fig:qualitative_pcbbank}
\end{figure*}

Another example from PCB-Bank~\cite{pcb_bank} is present in \cref{fig:qualitative_pcbbank}. The models are tested in two settings: zero-shot anomaly detection (green box) and anomaly detection with manufacturing process information (pink box). In the zero-shot anomaly detection setting, all models fail to correctly identify the missing button, although Triad detects a missing substrate that might be part of the design. This indicates a need for anomaly detection that incorporates information from the manufacturing process. 
The anomaly detection with manufacturing process information is presented in the pink boxes. 
In this example, Triad demonstrates its superior ability to utilize information from the PCB manufacturing process and precisely identify the missing button. 

\subsection{Failure case study}
\begin{figure*}[htbp]
  \centering
  \begin{subfigure}[t]{0.23\linewidth}
    \centering
    \includegraphics[width=\linewidth]{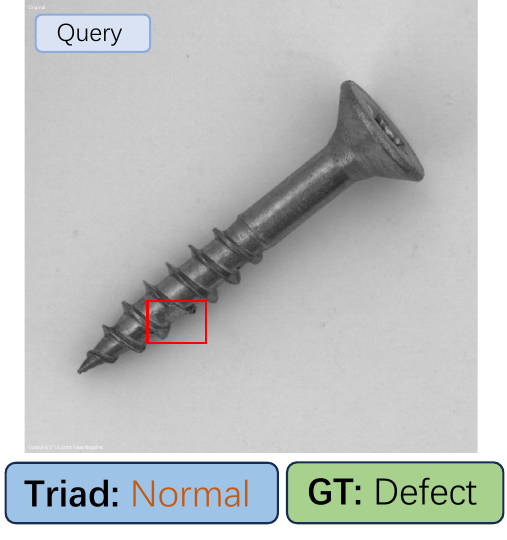} 
    \caption{Screw~(MVTec-AD)}
  \end{subfigure}
  \begin{subfigure}[t]{0.23\linewidth}
    \centering
    \includegraphics[width=\linewidth]{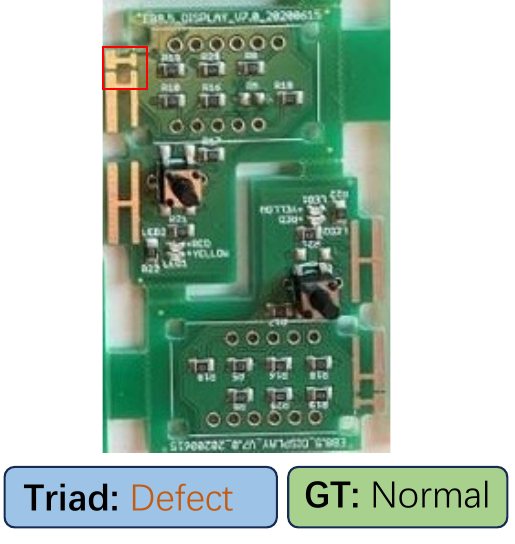}
    \caption{PCB7~(PCB-Bank)}
  \end{subfigure}
  \begin{subfigure}[t]{0.23\linewidth}
    \centering
    \includegraphics[width=\linewidth]{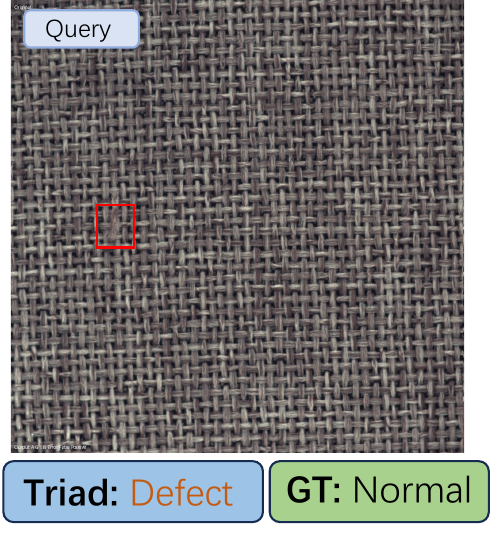}
    \caption{Carpet~(MVTec-AD)}
  \end{subfigure}
  \begin{subfigure}[t]{0.23\linewidth}
    \centering
    \includegraphics[width=\linewidth]{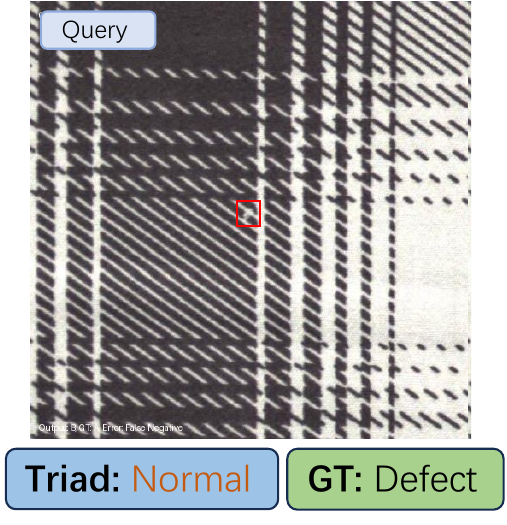}
    \caption{Grid cloth~(WFDD)}
  \end{subfigure}
  \caption{Failure cases from MVTec-AD, WFDD, and PCB-Bank.}
  \label{fig:failure_cases}
  \vskip -0.2in
\end{figure*}

\Cref{fig:failure_cases} presents four failure cases revealing Triad's current limitations: (a)~\textbf{Viewpoint ambiguity:} Requires 3D scans/X-ray imaging for better detection; (b) \textbf{Design-related errors:} Need CAD/GERBER design files for resolution; (c)~\textbf{Label-threshold conflict:} Defects below a prescribed size are labeled as normal by the dataset; (d) \textbf{Texture blindness:} Complex patterns obscure flaws in textiles, which necessitates texture-aware training data.

\section{More details for evaluation}
\label{sec: extra_eval}
Evaluation prompts are following templates shown in \cref{tab: prompt_temp}. For general LMMs and our method, questions and options are provided with additional instructions leading to a binary output of ``A'' or ``B'' for the convenience of calculating accuracy. For the LMM-based IAD methods, we keep their original prompts as much as possible. Specifically for AnomalyGPT~\cite{gu2024anomalygpt}, it has a hint before the question on MVTec AD~\cite{bergmann2019mvtec}.

The accuracy is calculated by retrieving a binary answer of ``Yes'' or ``No'' with keyword mapping. To evaluate the effect of context, the manufacturing process is directly added before the question. Some examples of the manufacturing process generated by LLM, which is used in evaluation on MVTec-AD are shown in \cref{tab: production_example}. 

\section{Task details}
\label{sec:task_details}
\begin{figure}
  \centering
  \includegraphics[width=\linewidth]{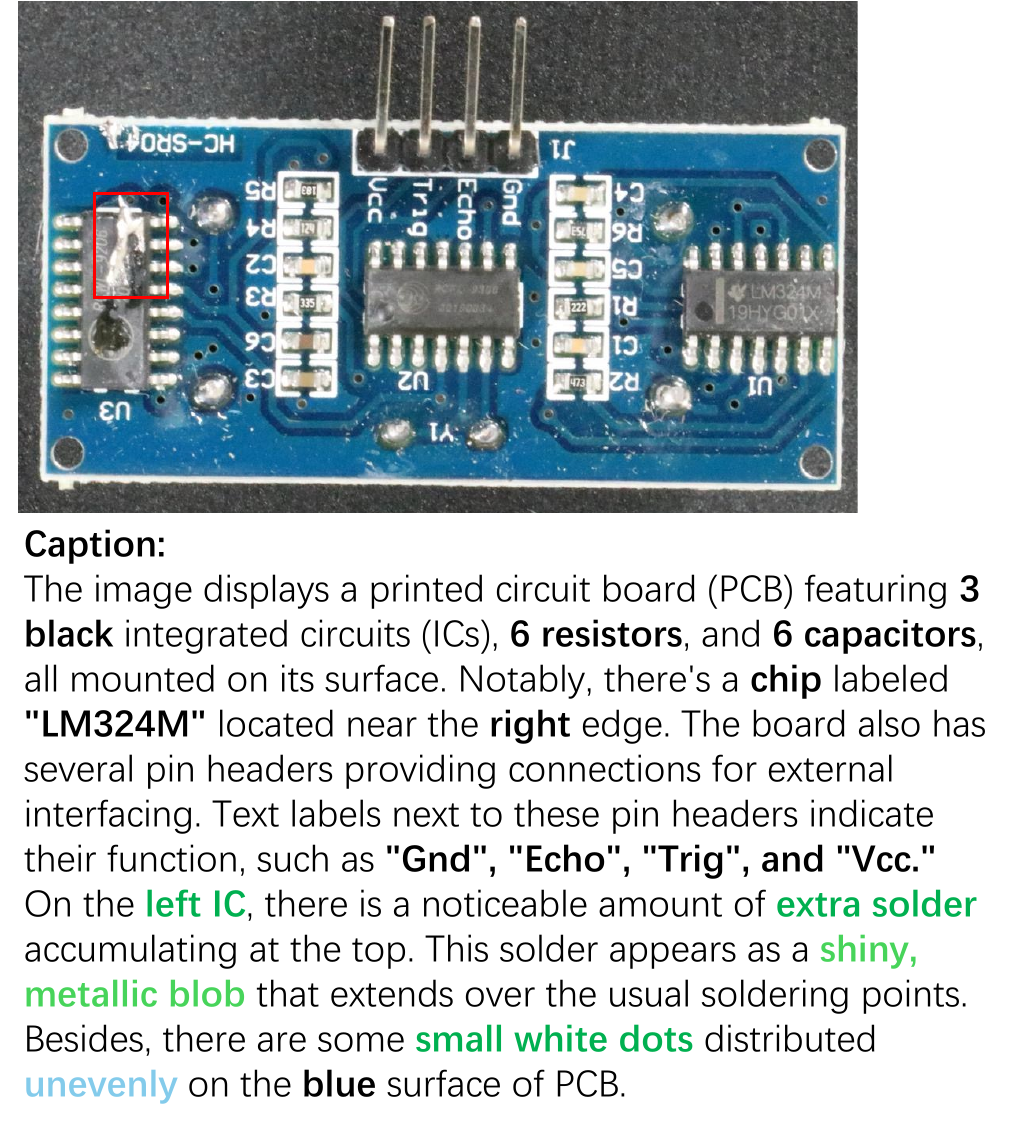}
  \vspace{-0.2in}
  \caption{An example of attribute-rich caption in InstructIAD}
  \label{fig:anno_example}
  \vspace{-0.05in}
\end{figure}

We have manually annotated detailed IAD captions on VisA~\cite{visa} and a part of RealIAD~\cite{realiad}. An example is shown in \cref{fig:anno_example} with a human-annotated caption. When labeling the IAD image, we first describe the product in the image and its detailed features following the general image-captioning task. Then, the defect is described by its type, location, texture, color, and shape. For multiple defects, each defect will be described separately with its relative position to the image. Later, captions are processed by Large Language Models~(LLMs) to construct different task-based instruction datasets.

\begin{figure*}
\centering
\begin{subfigure}[b]{0.49\linewidth}
    \centering
    \includegraphics[width=\linewidth]{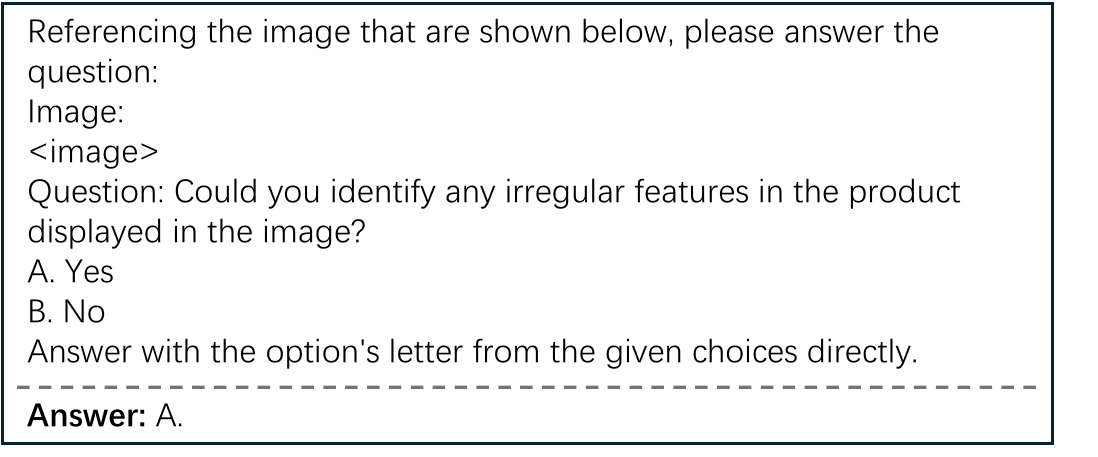}
    \caption{anomaly detection}
    \label{fig:task_a}
\end{subfigure}
\begin{subfigure}[b]{0.49\linewidth}
    \centering
    \includegraphics[width=\linewidth]{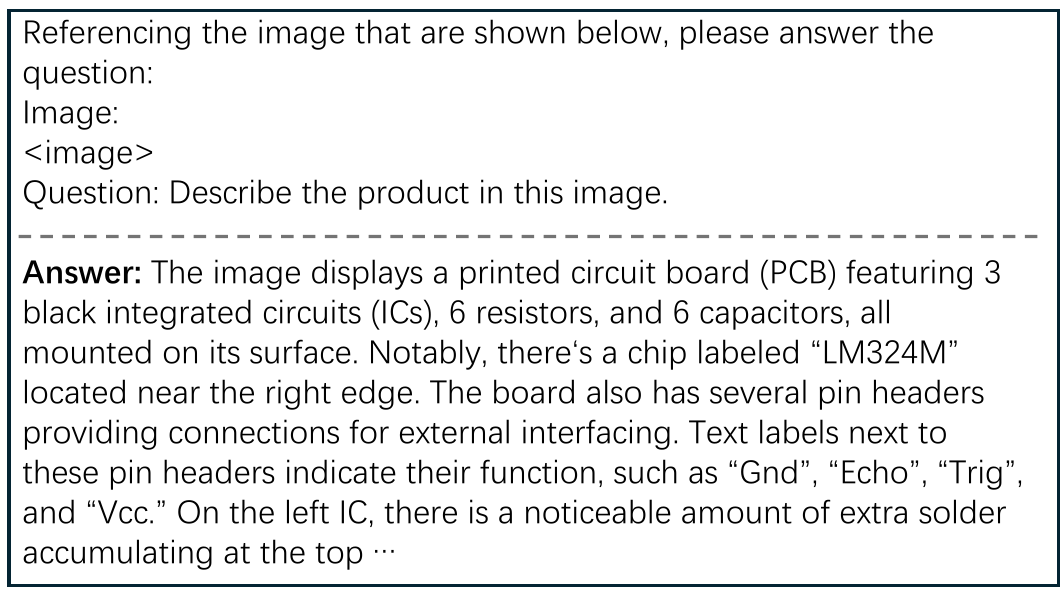}
    \caption{attribute-level caption}
    \label{fig:task_b}
\end{subfigure}

\begin{subfigure}[b]{0.49\linewidth}
    \centering
    \includegraphics[width=\linewidth]{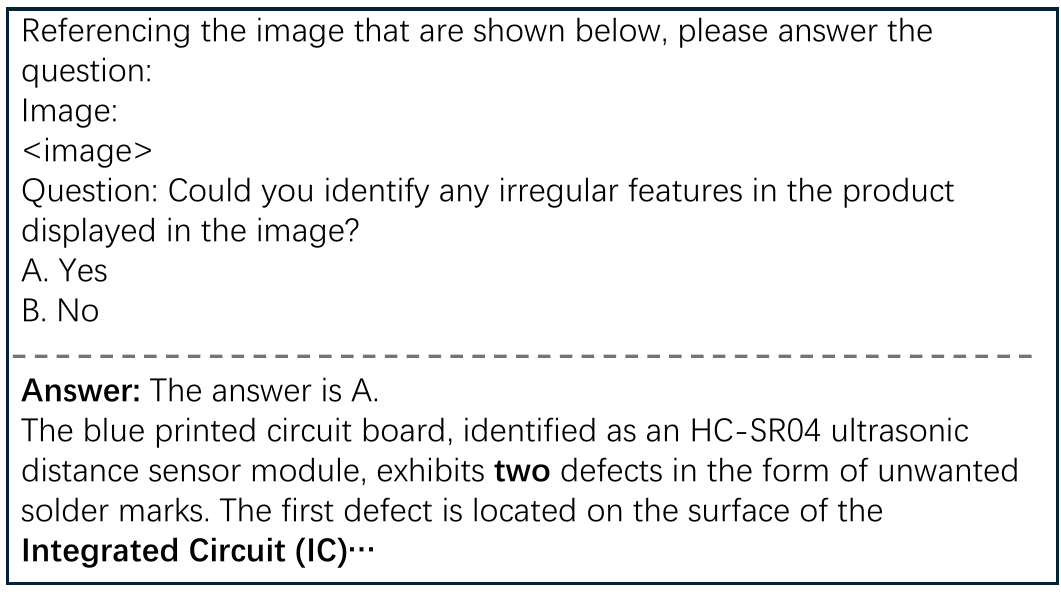}
    \caption{anomaly analysis}
    \label{fig:task_c}
\end{subfigure}
\begin{subfigure}[b]{0.49\linewidth}
    \centering
    \includegraphics[width=\linewidth]{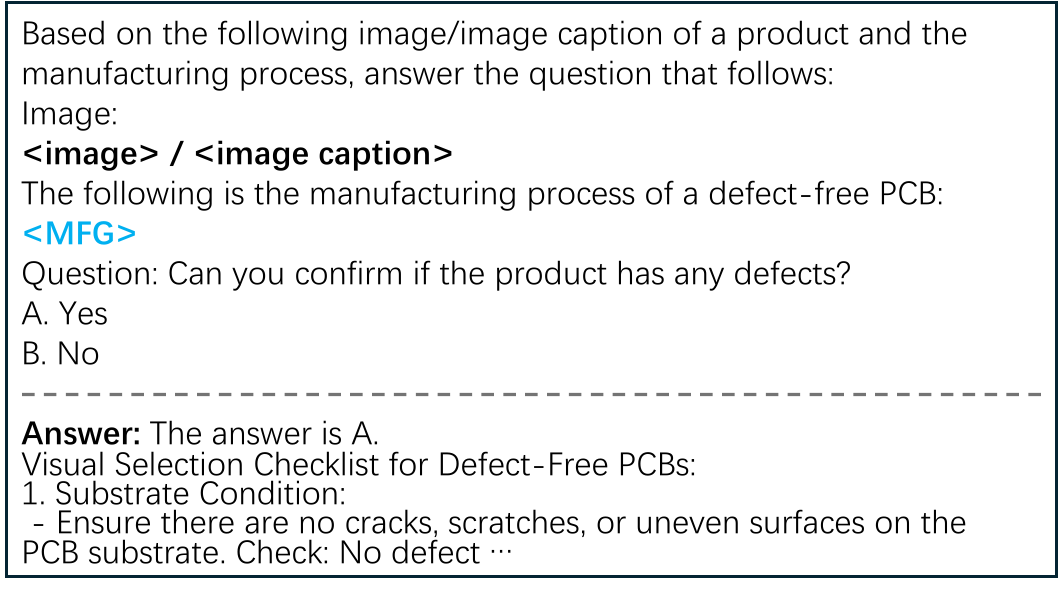}
    \caption{CoT-M data organization}
    \label{fig:task_d}
\end{subfigure}
\caption{Tasks templates used in InstructIAD and CoT-M}
\label{fig:task_details}
\vspace{-0.1in}
\end{figure*}

The examples of four tasks used in InstructIAD and CoT-M are shown in \cref{fig:task_details}. 

\section{Further results about EG-RoI}
\subsection{Binarized anomaly segmentation map}
Triad uses the binarized outputs of anomaly segmentation vision experts in EG-RoI. We observe that the range of anomaly scores varies between different vision experts, which makes it difficult to integrate them in a unified manner. The conventional normalization methods in ~\cite{chen2023april, cao2025adaclip, jeong2023winclip} are based on the entire test set, which is not feasible in real applications and requires a manually selected threshold for further binarization. Considering that vision experts only provide suspicious regions, we simply normalize the scores to the range 0-1 in one anomaly map and retain the scores that are larger than 0.9 on the anomaly map. As for threshold, \cref{tab:threshold_ablation} shows that although accuracy varies with different thresholds, Triad has robustness against the increase of false anomaly regions~(Pixel-FPR) and achieves a good performance when the True Positive Rate is low.

\subsection{Impact of the accuracy of anomaly maps}
We calculate the accuracy of anomaly maps provided by vision experts for different thresholds on the MVTec AD test set in \cref{tab: expert_acc}. Though selecting the best threshold, the performance of vision experts is typically poorer than Triad, and they also lack the ability to utilize extra context, such as the manufacturing process to improve their performance on new products.

\begin{table}
    \centering
    \caption{Comparison of 0-shot model and 1-shot model with or without CVM on Triad-lava-1.6.}
    \label{tab: cvm}
    \resizebox{\linewidth}{!}{
    \begin{tabular}{@{}c|cc@{}}
    \toprule
        model & 1-shot base & + MFG Proc.  \\ \midrule
         0-shot model & 85.0\% & 87.5\%~\stdvu{2.5\%} \\ 
        1-shot model & 87.8\% & 88.3\%~\stdvu{0.5\%} \\
        1-shot model~(CVM) & 87.7\% & 88.4\%~\stdvu{0.7\%}\\ 
        \bottomrule
    \end{tabular}}
    \vspace{-0.1in}
\end{table}

\begin{table*}
  \centering
  \small
  \renewcommand\arraystretch{1.0}
  \caption{Impact of threshold in EG-RoI (LLaVA-1.6/MuSc).-0.1 means the top 10\% regions with the lowest likelihood are marked, so does -0.2.}
  \vspace{-2mm}
  \resizebox{1.0\linewidth}{!}{
  \begin{tabular}{@{}c|ccccccccccc@{}}
  \toprule
  Threshold  & 
  0.9 & 
  0.8 &  
  0.7 &
  0.6 &
  0.5 &
  0.4 &
  0.3 &
  0.2 &
  0.1 &
  -0.1 &
  -0.2 \\ \midrule
  base &
  85.0\% &
  86.8\% &
  85.5\% &
  85.2\% &
  84.3\% &
  85.1\% &
  83.9\% &
  84.5\% &
  83.8\% &
  84.4\% &
  84.9\% \\
  +MFG (LLM) &
  87.5\% &
  87.8\% &
  86.7\% &
  86.5\% &
  86.1\% &
  85.8\% &
  85.2\% &
  85.3\% &
  84.6\% &
  84.9\% &
  84.7\% \\ 
  $\Delta$ &
  2.5\%{\color{ForestGreen}$\uparrow$} &
  1.0\%{\color{ForestGreen}$\uparrow$} &
  1.2\%{\color{ForestGreen}$\uparrow$} &
  1.3\%{\color{ForestGreen}$\uparrow$} &
  1.8\%{\color{ForestGreen}$\uparrow$} &
  0.7\%{\color{ForestGreen}$\uparrow$} &
  1.3\%{\color{ForestGreen}$\uparrow$} &
  0.8\%{\color{ForestGreen}$\uparrow$} &
  0.8\%{\color{ForestGreen}$\uparrow$} &
  0.5\%{\color{ForestGreen}$\uparrow$} &
  0.2\%{\color{red}$\downarrow$} \\\midrule
  Pix-TPR (Expert) &
  32.1\% &
  62.6\% &
  80.1\% &
  89.9\% &
  95.8\% &
  98.7\% &
  99.6\% &
  99.9\% &
  100.0\% &
  0.0\% &
  0.1\% \\
  Pix-FPR (Expert) &
  0.6\% &
  3.3\% &
  8.3\% &
  16.4\% &
  28.3\% &
  43.4\% &
  62.0\% &
  82.3\% &
  96.7\% &
  3.3\% &
  17.8\% \\
  \bottomrule
  \end{tabular}}
  \label{tab:threshold_ablation}
\end{table*}

\begin{table*}
\small
\caption{Accuracy calculated by the anomaly map we used on MVTec AD. The (0shot, 1shot) test is conducted on Triad-ov.}
\resizebox{1.0\linewidth}{!}{
\begin{tabular}{c|ccccccccc|c|cc}
\toprule
Threshold & 0.1 & 0.2 & 0.3 & 0.4 & 0.5 & 0.6 & 0.7 & 0.8 & 0.9 & Best Acc & Triad~(Base) & Triad~(+MFG)\\ \midrule
AnomalyClip~\cite{zhou2023anomalyclip} & 73.3\% & 75.1\% & 76.8\% & 77.4\% & 78.4\% & 79.5\% & 80.3\% & 80.2\% & \textbf{80.9\%} & 80.9\% & (90.5\%, 92.9\%) & (91.6\%, 93.0\%)\\
MuSc~\cite{li2024musc} & 72.3\% & 72.3\% & 72.3\% & 72.3\% & 72.3\% & \textbf{86.3\%} & 66.1\% & 39.5\% & 28.5\% & 86.3\% & (91.2\%, 92.9\%) & (92.6\%, 94.1\%)\\ 
\bottomrule
\end{tabular}}
\label{tab: expert_acc}
\vspace{-0.1in}
\end{table*}

\section{Other Explains}
\subsection{Confidence Voting Mechanism}
\label{sec:cvm}
Confidence Voting Mechanism~(CVM) is used in one-shot evaluation with the zero-shot model boosting the one-shot model in Traid~(LLaVA-1.6) due to the lack of multi-image ability of LLaVA-1.6~\cite{llavanext}. In LMMs, references do not always bring benefits; instead, they could be misleading if the reference has some differences from the query image. Thus, the opinion of the zero-shot model is worth thinking about whether LMM is not strong enough for understanding multiple images. Confidence Voting Mechanism uses scores to collect the results of zero-shot model and one-shot model and give a combined result.
As LMMs are bad at outputting confidence scores, the Confidence Voting Mechanism uses the option's word predict score from the language model and compares them by following rules: If two models reach a consensus, directly give the result; else if one model gives a negative opinion, which is saying the query is non-defective, its word predict score would be compared with the reference predict score by the same model. If the model is trusting the query more like a normal sample than the reference, it will be trusted and result in a non-defective judgment; otherwise, the opposite opinion will be adopted. The reason for comparing with the reference prediction score is that the space of predicted scores by the two models is not the same and unable to be normalized, so direct comparison makes no sense for the result. The effect of the Confidence Voting Mechanism is shown in \cref{tab: cvm}.

\begin{table*}
    \centering
    \caption{Prompt template used in evaluation. Manufacturing processes for each class of product are inputted as context.}
    \label{tab: prompt_temp}
    \begin{tabularx}{\textwidth}{XXX}
    \toprule
    Different methods & \multicolumn{1}{c}{0shot prompt} & \multicolumn{1}{c}{+MFG. proc prompt}\\ \midrule
    General LMMs\newline(Traid on LLaVA-1.6)
    & \textit{\textless image\textgreater} \newline Are there any defects on the \textit{\textless object\_name\textgreater} in this image? \newline A. Yes \newline B. No \newline Answer with the option's letter from the given choices directly. 
    & \textit{\textless image\textgreater} \newline \textit{\textless context\textgreater} \newline Are there any defects on the \textit{\textless object\_name\textgreater} in this image? \newline A. Yes \newline B. No \newline Answer with the option's letter from the given choices directly. \\ \midrule
    LLaVA-OneVision\newline(Triad on LLaVA-OneVision)
    & Referencing the image that is shown below, please answer the question:\newline Image:\newline\textit{\textless image\textgreater} \newline Question: Are there any defects on the \textit{\textless object\_name\textgreater} in this image? \newline A. Yes \newline B. No \newline Answer with the option's letter from the given choices directly. 
    & Referencing the image and production process that are shown below, please answer the question:\newline Image:\newline\textit{\textless image\textgreater} \newline \textit{\textless context\textgreater} \newline Question: Are there any defects on the \textit{\textless object\_name\textgreater} in this image? \newline A. Yes \newline B. No \newline Answer with the option's letter from the given choices directly. \\ \midrule
    Myriad & \textit{\textless Image\textgreater}\textit{\textless ImageHere\textgreater}\textit{\textless $\backslash$Image\textgreater}\newline This image may be simulated by photo editing. According to IAD expert opinions and corresponding visual descriptions, find out if there are defects in this image. & \textit{\textless Image\textgreater}\textit{\textless ImageHere\textgreater}\textit{\textless $\backslash$Image\textgreater}\newline \textit{\textless context\textgreater} \newline This image may be simulated by photo editing. According to IAD expert opinions and corresponding visual descriptions, find out if there are defects in this image. \\ \midrule
    AnomalyGPT & \textit{\textless /Img\textgreater}\newline \textit{\textless hint\textgreater} Is there any anomaly in the image? 
    & \textit{\textless /Img\textgreater}\newline\textit{\textless context\textgreater} \newline \textit{\textless hint\textgreater} Is there any anomaly in the image?\\ \bottomrule
    \end{tabularx}
\end{table*}

\begin{table*}
    \centering
    \caption{Examples of the manufacturing process generated by LLM about products in MVTec AD.}
    \label{tab: production_example}
    \begin{tabularx}{\textwidth}{cX}
    \toprule
    Class Name & \multicolumn{1}{c}{MFG Proc.} \\ \midrule
    Hazelnut
    & The following is the production process of the hazelnuts:\newline1. Harvesting: Hazelnuts are harvested from trees when they are ripe, usually in the late summer or early fall. Some variations in the size and color of the hazelnuts can occur based on the variety of tree and the growing conditions. This is not a defect, as it is normal for hazelnuts to have some variation in size and color.\newline2. Drying: After harvest, the hazelnuts are dried to reduce their moisture content. This helps to preserve the nuts and prevent spoilage. Some variation in the drying time and temperature can occur, which may cause slight differences in the color and texture of the hazelnuts. However, this is not a defect, as long as the hazelnuts have been dried to the correct moisture level.\newline3. Sorting: The dried hazelnuts are then sorted to remove any impurities, such as leaves, stems, and rocks. Some variation in the size and shape of the hazelnuts can occur during the sorting process. This is not a defect, as long as the hazelnuts are of a consistent size and shape suitable for the intended use.\newline4. Cracking: The sorted hazelnuts are then cracked to remove the shell. Some variation in the size and shape of the hazelnut kernels can occur during the cracking process. This is not a defect, as long as the kernels are intact and of a consistent size and shape.\newline5. Grading: The cracked hazelnuts are then graded based on their size and quality. Some variation in the size and quality of the kernels can occur. This is not a defect, as long as the hazelnuts meet the required standards for their intended use.\newline6. Packaging: The graded hazelnuts are then packaged for sale. Some variation in the packaging process can occur, which may cause slight differences in the appearance of the final product. However, this is not a defect, as long as the hazelnuts are properly protected and labeled.\\ \midrule
    Pill & The following is the production process of the pills:\newline1. Formulation: This is the first step where the active ingredients, excipients, and binders are mixed together in a specific ratio to form a powdered mixture. The differences that may occur in this step are related to the uniformity of the mixture, which can affect the dose of the active ingredient in each pill. However, these variations are usually not a defect as long as they fall within the acceptable range set by regulatory bodies.\newline2. Granulation: The powdered mixture is then moistened with a binder solution to form granules. The granules are then dried and screened to ensure uniformity in size. The differences that may arise in this step include variations in the size and shape of the granules, which can affect the flow properties of the material during compression. However, these differences are not considered defects as long as they do not significantly impact the uniformity or quality of the final product.\newline3. Compression: The dried granules are compressed into tablets using a tablet press. The differences that may occur in this step include variations in the weight, hardness, and thickness of the tablets, which can affect the dissolution rate and bioavailability of the active ingredient. However, these differences are not considered defects as long as they fall within the acceptable range set by regulatory bodies.\newline4. Coating: The compressed tablets are then coated with a film of polymer or sugar to improve their stability, appearance, and taste. The differences that may occur in this step include variations in the thickness and uniformity of the coating, which can affect the release profile of the active ingredient. However, these differences are not considered defects as long as they do not significantly impact the stability or efficacy of the final product. \\ 
    \bottomrule
    \end{tabularx}
\end{table*}

\begin{table*}
    \centering
    \caption{The manufacturing process of cables generated by LLM and GPT.}
    \label{tab:cable_mfg}
    \begin{tabularx}{\textwidth}{cX}
    \toprule
    Class Name & \multicolumn{1}{c}{MFG Proc.} \\ \midrule
    Cable~(GPT)
    & The following is the production process of the cables:\newline
1. Copper Wire Drawing: \newline
   ~~~~- Start with large copper rods and draw them through a series of progressively smaller dies to create fine strands of copper wire.\newline
2. Stranding:\newline
   ~~~~- Multiple fine copper wires are stranded together to form a single conductor for each wire.\newline
3. Insulation Extrusion:\newline
   ~~~~- Extrude insulation material around each stranded copper conductor. The insulation material is heated and then extruded around the wire as it passes through a die.\newline
4. Color Coding:\newline
   ~~~~- During the insulation process, the insulation is color-coded with different colors to differentiate the wires within the cable.\newline
5. Cable Assembly:\newline
   ~~~~- The three insulated wires are then twisted or laid together to form a single cable. This step ensures that the wires are held together in a compact and organized manner.\newline
6. Outer Sheath Extrusion:\newline
   ~~~~- An outer sheath is extruded around the assembled wires to provide additional protection and integrity to the cable.\newline
7. Cooling:\newline
   ~~~~- The extruded cable is passed through a cooling bath or air-cooling system to solidify the insulation and outer sheath materials.\\ \midrule
    Cable~(LLM) & The following is the production process of the cables:\newline1. Stripping: This is the process of removing the outer jacket of the cable to expose the inner conductors. Differences in the length or precision of the stripped area may occur, but they are not considered defects as long as they are within the tolerable range specified by industry standards.\newline2. Twisting: In this step, the conductors are twisted together to form a cable. The number of twists per inch and the tightness of the twist may vary slightly from one production to another, but these differences do not affect the functionality or quality of the cable.\newline3. Insulation: Each twisted pair of conductors is covered with an insulating material to prevent electrical interference. The thickness and uniformity of the insulation may differ slightly between productions, but these differences are not considered defects as long as they comply with industry standards.\newline4. Braiding: A braided shield is added around the insulated conductors to provide additional protection against electrical interference. The tightness and uniformity of the braid may differ slightly between productions, but these differences are not considered defects as long as they do not affect the cable's performance.\newline5. Jacketing: A final outer jacket is added to encase all the inner components of the cable. The thickness and uniformity of the jacket may differ slightly between productions, but these differences are not considered defects as long as they are within the tolerable range specified by industry standards.\\ 
    \bottomrule
    \end{tabularx}
\end{table*}
%
%

\end{document}